\documentclass{article}

\usepackage{PRIMEarxiv}

\usepackage[utf8]{inputenc} 
\usepackage[T1]{fontenc}    
\usepackage{url}            
\usepackage{booktabs}       
\usepackage{amsfonts}       
\usepackage{nicefrac}       
\usepackage{microtype}      
\usepackage{lipsum}
\usepackage{fancyhdr}       
\usepackage{graphicx}       
\usepackage{color}
\usepackage{lineno,hyperref}
\usepackage{harveyballs}
\usepackage{verbatim}
\usepackage{marvosym}
\usepackage{tikzsymbols}
\usepackage{IEEEtrantools}
\usepackage{comment}
\usepackage{graphicx}
\usepackage{lscape}
\usepackage{courier}
\usepackage{multirow}
\usepackage{amsmath}
\usepackage{listings}
\usepackage{courier}
\usepackage[square,numbers]{natbib}
\usepackage{amssymb}
\usepackage{algorithm}
\usepackage{algpseudocode}
\usepackage[normalem]{ulem}
\usepackage[normalem]{ulem}
\usepackage{latexsym}
\usepackage{longtable,geometry}
\usepackage{rotating}

\modulolinenumbers[5]

\modulolinenumbers[5]

\graphicspath{{media/}}     

\title{A Parametric Similarity Method: \\ Comparative Experiments based on \\ Semantically Annotated Large Datasets
\thanks{\textit{\underline{Accepted manuscript}}:\\ 
\textbf{Cite as:} Antonio De Nicola, Anna Formica, Michele Missikoff, Elaheh Pourabbas, Francesco Taglino,
A parametric similarity method: Comparative experiments based on semantically annotated large datasets, Journal of Web Semantics, Volume 76, April 2023, 100773, ISSN 1570-8268, https://doi.org/10.1016/j.websem.2023.100773.} 
}

\author{
  Antonio De~Nicola \\
  Italian National Agency for New Technologies, Energy and Sustainable Economic Development (ENEA)\\ Casaccia Research Centre\\
  Via Anguillarese 301\\ 
  I-00123, Rome\\
  Italy\\
  \texttt{antonio.denicola@enea.it} \\
   \And
  Anna Formica, Michele Missikoff, Elaheh Pourabbas, Francesco Taglino \\
  Institute of Systems Analysis and Computer Science ``Antonio Ruberti''\\
  National Research Council (IASI-CNR)\\
  Via dei Taurini 19\\
  I-00185, Rome\\
  Italy\\
\texttt{\{anna.formica, michele.missikoff, francesco.taglino\}@iasi.cnr.it} \\
}

\begin{document}
\maketitle

\begin{abstract}

We present the parametric method \textit{SemSim$^p$} aimed at measuring semantic similarity of digital resources. \textit{SemSim$^p$} is based on the notion of \textit{information content}, and it leverages a reference ontology and taxonomic reasoning, encompassing different approaches for weighting the concepts of the ontology. In particular, weights can be computed by considering either the available digital resources or the structure of the reference ontology of a given domain. 
\textit{SemSim$^p$} is assessed against six representative semantic similarity methods for comparing sets of concepts proposed in the literature, by carrying out an experimentation that includes both a statistical analysis and an expert judgement evaluation.
To the purpose of achieving a reliable assessment, we used a real-world large dataset based on the Digital Library of the Association for Computing Machinery (ACM), and a reference ontology derived from the ACM Computing Classification System (ACM-CCS). 
For each method, we considered two indicators.
The first concerns the degree of confidence to identify the similarity among the papers belonging to some special issues selected from the ACM Transactions on Information Systems journal, 
the second the Pearson correlation with human judgement. 
The results reveal that one of the configurations of \textit{SemSim$^p$} outperforms the other assessed methods. 
An additional experiment performed in the domain of physics shows that, in general, \textit{SemSim$^p$} provides better results than the other similarity methods.


\end{abstract}




\section{Introduction}
\label{sec:Introduction}

Similarity is a fundamental research topic relevant to different applications, from recommender systems \cite{10.1145/3372154}, text classification \cite{Liu2021}, health \cite{Abdelaziz2017}, to geographic information systems (GIS) \citep{9086014, prudhomme2020interpretation}, just to mention a few. In particular, semantic similarity relying on the use of knowledge-based techniques \cite{Hassanpour2014, Zhu2017, Bogdanovic2021, Formica2021} represents a promising research field.
Along this direction, we present the parametric semantic similarity method \textit{SemSim$^p$}. It is derived from \textit{SemSim} \citep{DFMPT19, FMPT13}, which has been conceived for evaluating semantic similarity of annotated resources (i.e., real world entities). There are no restrictions about the type of analyzed resources, they can be images, technical reports, descriptive brochures, and any other artifact. The only requirement is that a resource has to be semantically annotated, i.e., its content is described by a collection of concepts or features, referred to as a {\it semantic annotation vector} ({\it annotation vector} for short). Furthermore, such concepts cannot be freely chosen but they have to be selected from a {\it reference ontology}. 
In our work, the reference ontology is a taxonomy, i.e., a collection of concepts from a given application domain organized according to the ISA relationship (ISA hierarchy) \citep{FMPT13}. 
In order to evaluate the semantic similarity of digital resources, \textit{SemSim} requires a \textit{weighted} reference ontology, where each concept is associated with a weight.
Such a weight represents the specificity (or, inversely, the genericity) of the concept in the given domain that is inversely related to the cardinality of the set of the described resources (i.e., its extension) in the given domain.
\textit{SemSim} evaluates the semantic similarity of digital resources by exploiting the structure of the ontology. In particular, it
computes the semantic similarity of pairs of annotation vectors on the basis of the similarity between concepts from the ontology, by relying on the {\it information content} approach \citep{Lin}. 
\textit{SemSim} has been assessed with experiments in several case studies \citep{DFMPT19, FMPT13, FMPT16}, showing its efficiency and better performances with respect to well-known methods in the literature.

In this paper, we present the parametric method \textit{SemSim$^p$}, which is an evolution of \textit{SemSim}. It essentially depends on two parameters: the method used for computing the weights associated with the concepts of the ontology, and a normalization factor
 adopted when the two compared annotation vectors have different cardinalities.
The objective is the validation of \textit{SemSim$^p$} by means of a comparative assessment, showing 
that
\textit{SemSim$^p$} configured with a new selection of parameters
improves the performance of \textit{SemSim}, as well as it outperforms some of the most representative semantic similarity methods proposed in the literature.
In particular, we assessed \textit{SemSim$^p$} by contrasting it against six of the most popular similarity methods for comparing annotation vectors, organized in two groups.
One group, containing Dice \citep{Dice1945}, Jaccard \citep{Jaccard1912}, and Sigmoid \citep{LIKAVEC2019}, relies on set-theoretic methods, since the similarity scores are derived by applying set-theoretic operations on the annotation vectors associated with the digital resources.
The second group, formed by WNSim \citep{Shajalal2019}, and the methods proposed by Rezaei \& Fr{\"a}nti \cite{Reza2014} and Haase et al. \cite{Hasse2004}, includes taxonomy-based methods, which require pairs of concepts in the annotation vectors to be compared.

Evaluating a similarity reasoning method is not a simple task, it presents two orders of difficulties: one relies on the choice of the set of resources (dataset) to be analyzed, and the other is the actual benchmark, i.e., the reference against which it is possible to measure the performance of the selected method. With regard to benchmarking techniques, the most used one is based on human judgement \citep{Mandeep2001} \cite{TOCH201116}. In essence, considering a set of resources with their annotation vectors, a group of people is asked to pairwise match and assign them a similarity score. However, it is well-known that the human decision about similarity can be subjective, depending on the perspective adopted in looking at the resources, the relevant features, the purpose in mind and, finally yet importantly, the selected context. Secondly, a robust experiment requires a statistically relevant number of resources to be addressed. However, the greater the number of resources the harder the effort of people to perform the evaluation task.

We performed two different experiments based on the articles published in the Digital Library of the Association for Computing Machinery (ACM-DL). Since it is unrealistic to ask a group of people to pairwise check thousands of annotation vectors, producing millions of similarity scores, the key idea of this experimentation is to use some ACM special issues as benchmarks, because they contain papers whose semantic similarity, on average, is supposed to be greater than the one of a collection of papers randomly selected. In fact, such papers are gathered by the editor according to the research topic indicated in the call for paper. In particular, the first experiment is based on a statistical analysis without human intervention, and the second one on expert judgement (EJ), limited to the ACM special issues. The results of both the experiments reveal that one of the configurations of \textit{SemSim$^p$} outperforms the other assessed methods. 
In order to confirm these results, a further experiment was performed by relying on the articles published by the American Physical Society (APS). 
This additional experiment shows that, in general, \textit{SemSim$^p$} provides better results with respect to the selected similarity methods.

The paper is organized as follows. In the next section, the related work is introduced. 
In Section \ref{sect:semsimp} the parametric \textit{SemSim$^p$} method is presented. In Section \ref{sect:case-study}, the experiments are described, starting with the context description and the related research questions. 
In Section \ref{sect:discussion} the experimental results are discussed, and in Section \ref{sect:conclusion} the conclusion is presented.\\

\section{Related Work} \label{sect:related-work}

\begin{figure}
	\centering
	\includegraphics[scale=.75]{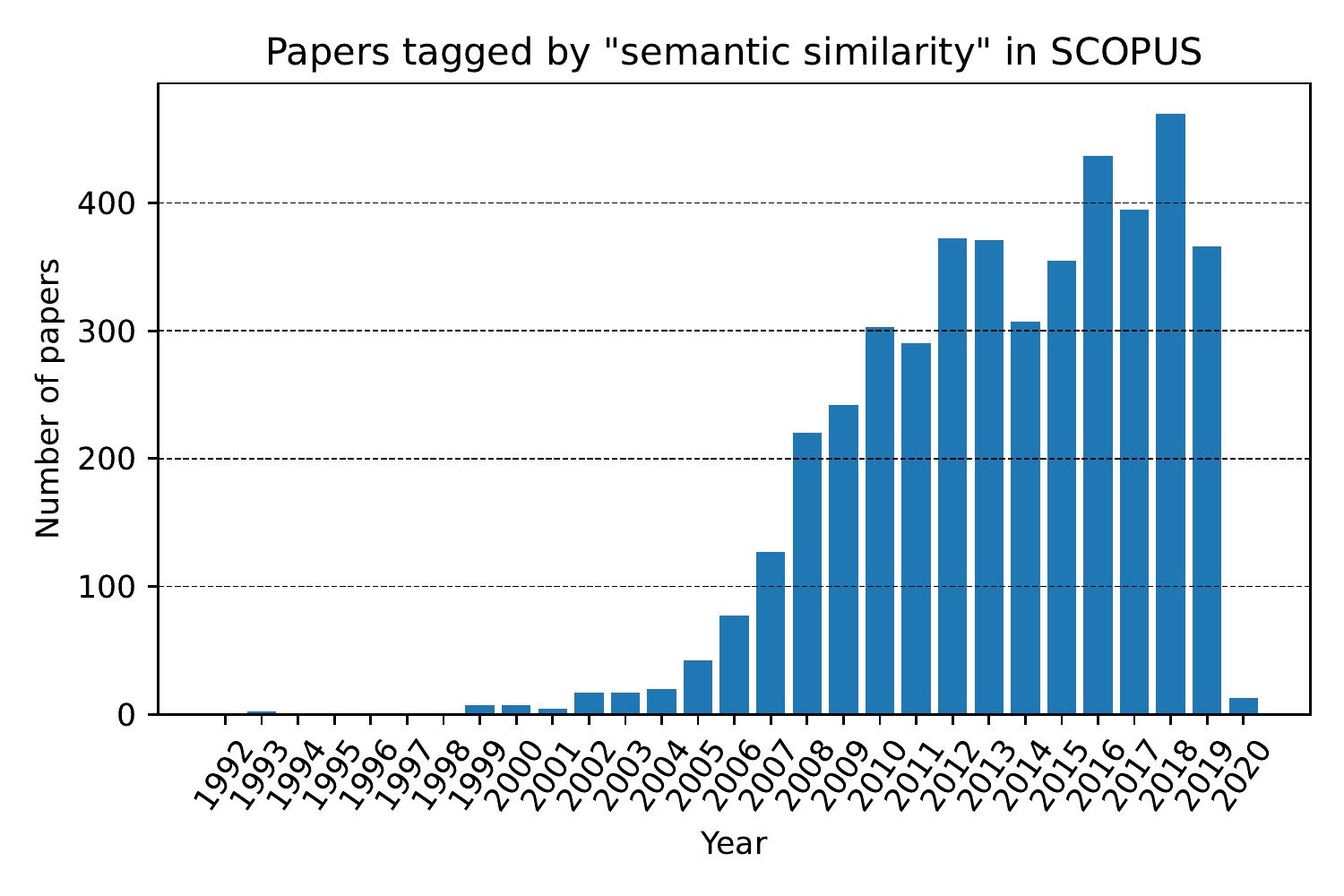}
	\caption{Histogram of all the papers indexed by SCOPUS  with ``semantic similarity'' (retrieval date: Nov, 2019).}\label{fig:InterestHistogram}
\end{figure}

Nowadays, we are witnessing an increasing interest of the research community on the topic of semantic similarity, both at theoretical and practical levels. In fact, it encompasses a variety of fields spanning from artificial intelligence, social sciences to medicine, genetics, etc.
In order to better emphasize the role of the semantic similarity research topic, we assessed its impact on the literature by considering the SCOPUS abstract and citation database. 
In this context, we ran a query\footnote{The query was run on the 19\textsuperscript{th} November, 2019.} aimed at retrieving all the papers tagged by the keyword ``semantic similarity'' in the time interval from 1992 to 2019, and  we retrieved 4,463 papers.  Figure \ref{fig:InterestHistogram} shows the distribution per year of the obtained papers,  where their increment in the considered time interval is evident. 
As we observe, more than 200 papers have been tagged by year in the interval from 2008 to 2019, and the highest number of papers, that is about 500, is related to 2018. 
Each paper is also associated with one or more  sectors that are shown in Table \ref{tab:SectorsOccurrences} in Annex A according to a descending order of number of papers.

\begin{figure}
	\centering
	\includegraphics[scale=.35]{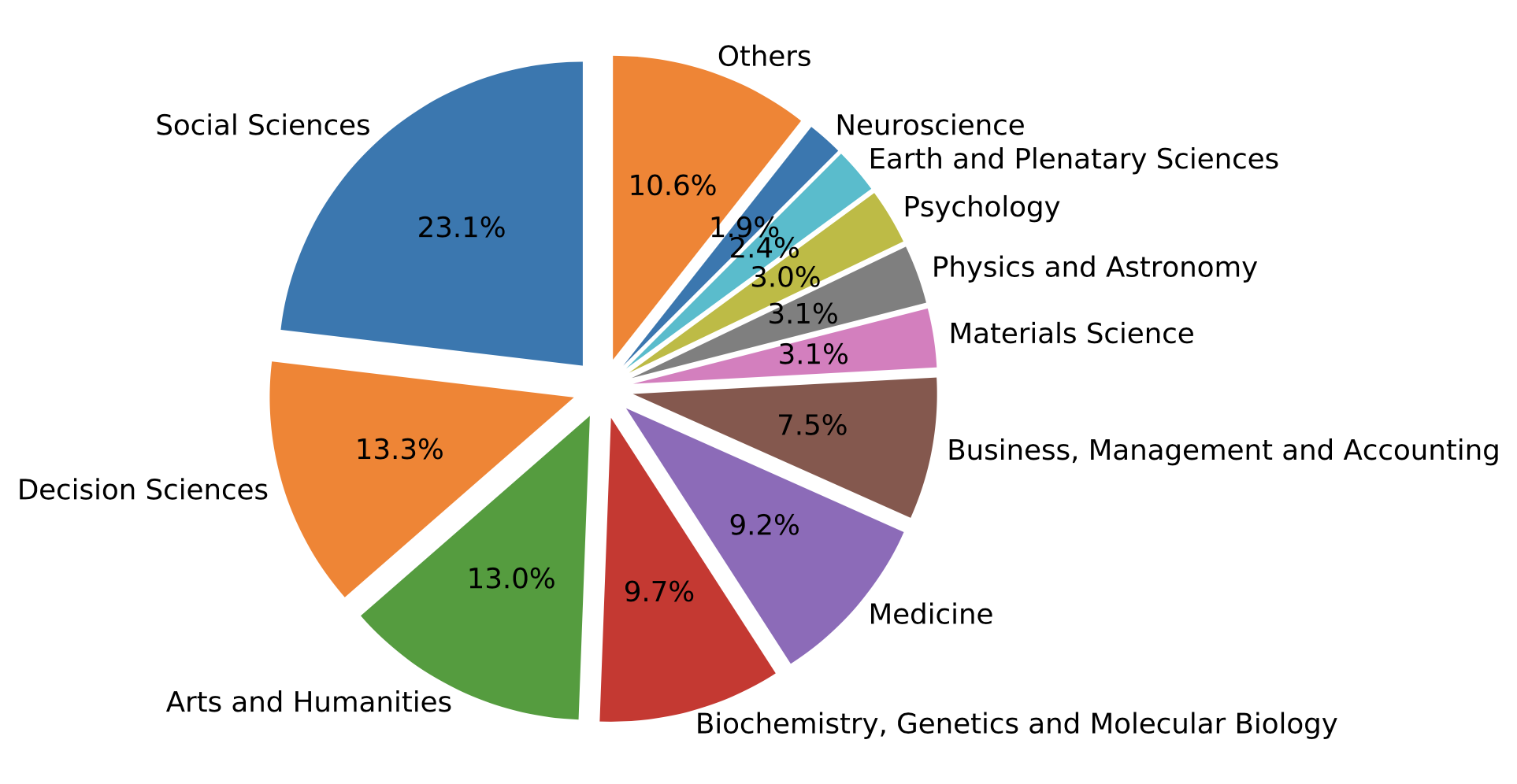}
	\caption{Percentage of papers indexed by SCOPUS with ``semantic similarity''  by application sectors.}\label{fig:impact}
\end{figure}

Among the retrieved sectors, we distinguished the {\it technical} sectors from the {\it application} ones. In the former, the papers likely present technical details, for instance, algorithms to compute semantic similarity or implementations of semantic similarity solutions. The latter,  i.e.,  the application ones, are illustrated in Figure \ref{fig:impact}. They concern social sciences, decision sciences, arts and humanities, biochemistry, genetics and molecular biology, medicine, etc. Among these sectors, {\it Social Sciences}, {\it Decision Sciences}, and {\it Arts and Humanities} involve a relevant high number of papers tagged with ``semantic similarity'', which are $23.3\%$,  $13.3\%$, and $13.0\%$ papers, respectively.

From a methodological point of view, methods for evaluating similarity can be categorized as follows  \cite{Chandra2021}: \textit{knowledge-based} \cite{Hassanpour2014, Zhu2017, Bogdanovic2021, Formica2021} (the focus of this paper), \textit{corpus-based} \citep{YANG2020100578} that can be further classified according to kernel-based models \cite{DAmato2007, Bloehdorn2007}, \textit{deep neural network-based} \cite{Huy2019, Samira2021}, and \textit{hybrid methods} \cite{Hassan2019}.
Currently, the focus of research is shifting towards deep neural network-based methods. However, these methods require high computational resources and lack interpretability, and the need for determining a balance between computational efficiency and performance is still a challenge \cite{Chandra2021}. With respect to these methods, in this work, we rely on a knowledge-based approach that, as illustrated in this paper, shows good performances and computational efficiency.

This section gives an overview of the existing works from the literature addressing the main  topics of this paper. Indeed, it is organized as follows: Section \ref{sec:OWM} focuses on the methods for weighting concepts in an ontology, Section \ref{sec:SimSetsOfConcepts} recalls some methods for computing similarity between sets of concepts and, in particular, the ones selected for comparison with $SemSim^p$ in our experiment, and
Section \ref{SemSimpApproachEvaluation} is about the approaches for evaluating them.

\subsection{Ontology Weighting Methods}\label{sec:OWM}

Extensional methods, also referred to as distributional \citep{Sanchez2011}, in general compute the information content of a concept on the basis of the frequency distribution of the corresponding terms in a corpus of text documents. Therefore, these methods leverage on the extensional semantics of the concepts since their probability can be calculated by considering the number of occurrences of each  concept in the text corpora. Extensional approaches were used by Jiang \& Conrath \cite{Jiang1997}, Lin \cite{Lin}, and Resnik \cite{Resnik1995} to estimate semantic similarity between concepts.
They also encompass the inverse document frequency (IDF) method, and the combination of term frequency (TF) and IDF \citep{Manning2008, tfidf}. In our work, we derived the concept frequency method and the annotation frequency method, respectively, from the one used by Resnik \cite{Resnik1995} and IDF (see Section \ref{sect:semsimp}). 

  The intensional methods, also referred to as intrinsic \citep{Sanchez2011}, compute the information contents of concepts on the basis of the conceptual relationships existing between them and mainly, on the basis of their taxonomic structure. With this regard, the method of Seco et al. \cite{Seco2004} is based on the number of concepts' hyponyms and the total number of concepts in the taxonomy. An extension of the mentioned method is presented by Meng et al. \cite{Zhou2012}, where also the depth of concepts in the taxonomy, i.e., the degree of generality of concepts, is considered.
 In \citep{Sanchez2011}, the authors claim that the taxonomical leaves are enough to describe and differentiate two concepts because {\it ad-hoc abstractions} (e.g., abstract entities) rarely appear in the universe of  discourse, but have an impact on the size of the taxonomy.
    In order to estimate the weight of a concept, Hayuhardhika et al. \citep{Hayuhardhika2013} propose to use the density factor of a concept, which is based on the sum of incoming and outgoing connections with other concepts versus the total number of connections in the ontology. 
    Finally, Abioui et al. \cite{Abioui2018} consider both the taxonomic structure and the other ontological relationships to derive concepts' weights. In our paper, we focused on taxonomies (i.e., structures organized according to ISA relationships) because, in general, they are used for classification purposes by actual communities (e.g., the ACM and APS communities).
  
With respect to the previous proposals, in our work both  the extensional and the intensional approaches have been addressed and experimented. 

    
\subsection{Similarity between Sets of Concepts}\label{sec:SimSetsOfConcepts}

In this section, since we represent a resource as a collection of concepts, we recall some proposals concerning the similarity between sets of concepts. In particular, we start from the approaches that were selected for the experiments conducted in this paper.

Traditionally, in order to evaluate the similarity between sets of concepts (features), the Dice \citep{Dice1945} and Jaccard \citep{Jaccard1912} measures were adopted, which can also be formulated according to the well-known Tversky model \citep{Tversky1977}. Furthermore, the Sigmoid similarity measure \citep{LIKAVEC2019}, which is an improvement of Dice, was also considered. 
As mentioned in the Introduction, these are the three set-theoretic methods that were considered in our experimentation, as formally recalled in Section \ref{sec:SelectedSimilarityMethodsForComparison}. The formal definitions of these three methods are provided in Table \ref{tab:SetTheoreticMethods} in Annex C.

With regard to the methods relying also on a taxonomy,
 we recall three similarity measures, the ones introduced by Rezaei \& Fr{\"a}nti \citep{Reza2014}, and Haase et al. \citep{Hasse2004}, and the WNSim similarity \citep{Shajalal2019}, that were also used in order to evaluate $SemSim^p$.
 They are formally given in Table \ref{tab:TaxonomyBasedMethods} in Annex C.
In particular, 
in \citep{Reza2014}, a similarity measure between sets of keywords was presented, which is based on matching the individual concepts of two groups by applying the well-known Wu and Palmer measure \citep{WuPalmer}, by using the WordNet taxonomy. 
In \citep{Hasse2004}, in order to compute the similarity of pairs of concepts belonging to different sets, the authors use the edge-based similarity measure proposed by Li et al. \citep{Li2006}, which combines the shortest path lengths and the depths of subsumers in the taxonomy non-linearly. 
In \citep{Shajalal2019}, the authors propose WNSim, a method for computing similarity between sets of keywords representing sentences in documents, by leveraging the Leacock and Chodorow similarity between concepts \citep{Leacock1998}. 
Both the methods proposed, respectively, by Haase et al. \citep{Hasse2004} and Shajalal \& Aono \citep{Shajalal2019} are asymmetric since, in comparing the two sets, the selection of the first one is relevant. Note that, as opposed to the mentioned methods, $SemSim^p$ is symmetric.


Among the other proposals that are based on the taxonomy, it is worth mentioning the paper of Cord{\`{\i} et al. \citep{CordiLMM05}, where the problem of defining a similarity metric between sets of concepts belonging to the same ontology is addressed. Essentially, two sets of concepts, named as {\it source} and {\it target} in the mentioned paper,  have a similarity degree equal to 1 if all the concepts in the source are in the target, up to a normalization factor. However, the similarity metric proposed by these authors is based on the Dijkstra's algorithm, i.e., on the evaluation of the minimal path between nodes in the graph. 
 Analogously  to the approach of Haase et al. \citep{Hasse2004}, concepts which are closer according to their positions in the taxonomy are semantically more similar than concepts that have a greater distance. 
Chen et al. \citep{Chen201719} present a semantic similarity measure for web service discovery. It can be applied to both textual descriptions and interface signatures. 
However, these two proposals are based on an enriched notion of semantic distance which relies on weighted paths combining different relationships of an ontology (e.g., is-a and has-a), and therefore we did not address them in our experimentation that leverages the information content of an ISA hierarchy.

Finally, in the paper of Jia et al. \citep{JiaLDL19}, semantic similarity has been investigated in the clinical context, in order to measure patient disease similarity. In the paper, the authors recall the most representative metrics proposed in the literature for evaluating the similarity between sets of concepts, including Dice, Jaccard, and  the information content approach.
However, the authors state that the choice of the most appropriate algorithm under various 
clinical scenarios is still a challenge, especially when the sizes of the sets to be compared are 
large or unbalanced, and they claim the need for further investigations about this research topic.


\subsection{Approaches to Semantic Similarity Evaluation}\label{SemSimpApproachEvaluation}
Traditionally, semantic similarity between concepts or words is assessed by developing experiments where people are asked to express a similarity value between terms. In some cases, such experiments bring to the definition of the so called \textit{golden datasets}. In this context, a golden dataset is defined as a list of pairs of terms, each associated with a numerical value representing the semantic similarity between the terms as derived from the human judgement evaluation. 
The use of a golden dataset is important since it can be employed as a benchmark in the evaluation and comparison of different similarity methods.
Among the existing golden datasets we recall:
\textbf{R\&G}, which is composed of 65 pairs of words representing general concepts evaluated by 51 college undergraduates  according to a scale from 0 to 4, where the higher the semantic similarity, the higher the score \citep{Rubenstein1965};
\textbf{M\&C}\footnote{\url{https://github.com/alexanderpanchenko/sim-eval/blob/master/datasets/mc.csv}} is composed of 30 pairs of words, which represent a subset of  R\&G evaluated by 38 persons \citep{Miller1991};
\textbf{R122} is composed of 122 pairs of words representing generic concepts evaluated by 14 up to 22 people, out of a total of 92 participants \citep{Szumlanski2013}.

In our case, we could not use one of the available golden datasets. In fact, there are no golden datasets including similarity scores for all possible pairs of concepts in the set and, in particular, concerning the ACM-CCS keywords, which is the domain of our experimentation.

The statistical method used for the experimentation analysis presented in our paper is similar to the approach used by K{\"o}hler et al. \cite{kohler2009clinical} to identify candidate diseases that best explain a set of clinical features. In fact, they  propose a statistical model to assign a \textit{p-value}\footnote{According to Mertens and Recker \cite{mertens2019new}, the \textit{p-value} indicates the probability of obtaining the
observed result or anything more extreme than that actually observed in the available sample data.} to the similarity score obtained by searching \textit{n} terms, corresponding to the probability of having a similarity score higher than, or equal to, the one obtained by randomly sampling with the Monte Carlo method \citep{kalos2009monte} the same number of query terms. 
If the highest-scoring candidate disease has a significant \textit{p-value}, this indicates that the syndrome is highly likely. Otherwise, the clinical features are not specific enough to allow a diagnosis.
More generally, De Nicola \& D'Agostino \cite{DeNicola2020} analyzed a semantic social network \cite{MIKA20075} and used a statistical approach based on \textit{p-value} to assess the diversity of groups of people with regard to some features characterizing them, as for instance performance indicators or some others related to psychological aspects.
With respect to these works where the diversity of groups is not an assumption, we changed the perspective as we assumed that there exist samples of similar digital resources (i.e, the special issues) that can be considered as a benchmark, and we used a statistical approach based on  \textit{t-value} \citep{bluman2009elementary} (see Section \ref{SemanticCohesion}) in order to assess the best method to evaluate semantic similarity.

\section{The Parametric \textit{SemSim$^p$} Method}
\label{sect:semsimp}

In this section, we present the parametric semantic similarity method  \textit{SemSim$^p$}, which is based on \textit{SemSim} introduced in \citep{FMPT13}. The \textit{SemSim} method has been revised from two points of view. One takes into account the methods conceived to assign weights to concepts.
The other one, instead, refers to a normalization factor embedded in the method, which is a coefficient that captures different counts of the annotation vectors cardinality.
Such a coefficient allows the similarity measures to be normalized to values between 0 and 1.
To this end, we recall some basic notions already introduced in \cite{FMPT13, FMPT16}. Then, we describe the  \textit{SemSim$^p$} method, and the different values that the normalization factor can assume. Successively, we discuss 
some methods to assign weights to concepts in order to calculate their semantic similarity.

\subsection{Basic Notions} \label{sect:basic-notion}

According to the definitions introduced by \cite{FMPT13, FMPT16}, an ontology $Ont$ is a taxonomy defined by the pair:
\begin{equation} 
	Ont = <\mathcal{C}, \textit{ISA}>
\end{equation}  
\noindent where ${\mathcal{C}}=\{c_i\}$ is a set of concepts, and \textit{ISA} is the set of pairs of concepts in ${\mathcal{C}}$ that are in subsumption ($\sqsubseteq$) relationship:
\begin{equation}  
	ISA = \{(c_{i},c_{j}) \in {\mathcal{C}}\times {\mathcal{C}} | c_{i} \sqsubseteq c_{j}\}
\end{equation}
where $c_{i} \sqsubseteq c_{j}$ means that $c_{i}$ is a child of $c_{j}$ in the taxonomy. 
Note that, in this work, we focus on taxonomies that are trees (i.e., tree-shaped taxonomies). 
A {\it Weighted Reference Ontology} ($WRO$) is then defined as follows:

\begin{equation} 
	WRO = <Ont, w>
\end{equation} 

\noindent where $w$, denoting the concept weighting function, is a probability distribution defined  on ${\mathcal{C}}$, such that given $c \in {\mathcal{C}}$, $w(c)$ is a decimal number in $[0,1]$. A simple tree-shaped taxonomy is introduced in Figure \ref{fig:small-ont}. It will be used in Section \ref{sect:weighting} in order to present the different ontology weighting methods.

\begin{figure}
	\centering
	\includegraphics[width=7cm, height=5cm]{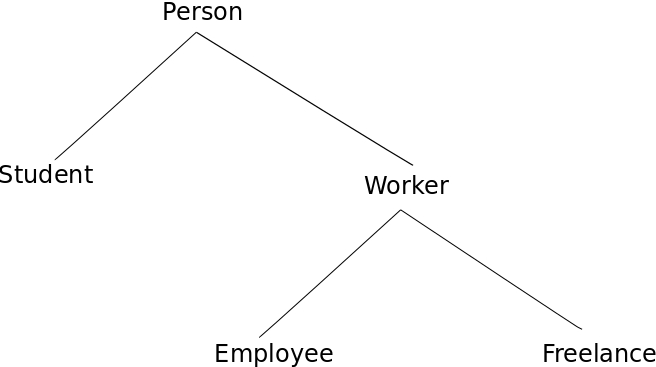}
	\caption{A simple taxonomy.}\label{fig:small-ont}
\end{figure}

Using the $WRO$, each resource is annotated by means of a semantic \textit{annotation vector}. An annotation vector $av$ is a collection of concepts of the ontology $Ont$, introduced to capture the semantic content of the corresponding resource. It is represented as follows:
\begin{equation} 
	av = (c_1,...,c_n),  
	c_i \in {\mathcal{C}}, i = 1,...,n.
\end{equation}

\subsection{The \textit{SemSim$^p$} algorithm} \label{sect:semsim-algo}


 The \textit{SemSim$^p$} method has been defined in order to calculate a semantic similarity degree between annotation vectors \citep{FMPT13}.
This method is based on two parametric functions, $consim_h$ (see Eq. (\ref{eq:consim})) and $semsim_{h, \mu}$ (see Eq. (\ref{eq:semsim})), the former used to compute the similarity of pairs of concepts (or features), whereas the latter conceived to calculate the similarity of pairs of annotation vectors.

In particular, given two concepts $c_{1}$, $c_{2}$, the similarity between them is given as follows:

\begin{equation} \label{eq:consim}
	consim_h(c_{1},c_{2})= \displaystyle \frac{2\times
		IC_h(lcs(c_{1},c_{2}))}{IC_h{(c_{1})}+IC_h(c_{2})} \\
\end{equation}

\noindent where {\it lcs}$(c_{1},c_{2})$ is the \textit{least common subsumer} of the concepts \textit{c}$_1$ and \textit{c}$_2$ in the taxonomy, i.e., 
 the least abstract concept of the ontology that subsumes both of them and, for any concept \textit{c} $\in$ ${\mathcal{C}}$, $IC_h(c)$ is defined as follows: 
 
\begin{equation}
IC_h(c)=
\begin{cases}
- log({\it w_h}(c)) &  \text{if   } h=\{CF, AF, TD\}\\
{\it iic(c)} & \text{if   }  h=\{IIC\} 
\end{cases}
\end{equation}


\noindent where  \textit{CF} ({\it Concept Frequency}), \textit{AF} ({\it Annotation Frequency}), \textit{TD} ({\it Top-Down topology-based}), and \textit{IIC} ({\it Intrinsic Information Content}) are ontology weighting  methods  that will be presented in the next section. Note that $IC_h(c)$, in the case  $h = \{CF,AF,TD\}$, is the information content of the concept $c$ as introduced in \cite{Lin}, whereas in the case  $h =\{IIC\}$,
it is defined according to \cite{Seco2004}.

Now, let us consider two annotation vectors,
say $av_1= (c_{11},\ldots, c_{1n})$ and $ av_2=(c_{21},\ldots, c_{2m})$, respectively. 
The $semsim_{h,\mu}$ function computes the $consim_h$ for each pair of concepts belonging to the set S = $av_1 \times av_2$, i.e., the Cartesian product of $av_1$, and $av_2$. 

We focus on the pairs that show high
affinity, and to this end, 
in line with the {\it maximum weighted matching} problem in bipartite graphs \citep{Dulmage}, 
a concept belongs to at most one pair.
Accordingly, 
$\mathcal P$$(av_1,av_2)$ is the set of sets of pairs, defined as follows:

	

\begin{equation}
\begin{array}{c}
\mathcal{P}(av_1,av_2) =  \{P \subset S | \hspace{0.1in}
\forall (\textit{c}_{1i}, \textit{c}_{2j}), (\textit{c}_{1q}, \textit{c}_{2k}) \in P, \\
\hspace{0.7in} \textit{c}_{1i} \neq \textit{c}_{1q},
\textit{c}_{2j} \neq \textit{c}_{2k}, |P| = min\{n, m\}\} \hspace{0.7in}
\end{array}
\end{equation}

The $semsim_{h, \mu}$ function identifies the set of pairs of concepts of the $av_1$ and $av_2$
that maximizes the sum of the $consim_h$ values. Formally, we have:

\begin{equation} \label{eq:semsim}
semsim_{h,\mu}(av_1, av_2)=
\displaystyle \frac{\max\limits_{P \in {\mathcal P}(av_1, av_2)}\big\{\sum\limits_{(\textit{c}_{1i},\textit{c}_{2j}) \in P}consim_h(\textit{c}_{1i},\textit{c}_{2j})\big\}}
{\mu(n,m)}
\end{equation}

\noindent where $\mu$, referred to as \textit{similarity normalization factor}, is defined as follows:

\begin{equation}
\mu(n,m) = 
\begin{cases}
{\it max(n,m)}\\
{\it min(n,m)}\\
{\it ave(n,m)}   & (\text{{\it arithmetic} {\it average(n,m)}} = \frac{n+m}{2}) \\
{\it gav(n,m)}   & (\text{{\it geometric} {\it average(n,m)}} = \sqrt{nm})
\end{cases}
\end{equation}

Note that in Eq. (\ref{eq:semsim}) the semantic similarity between annotation vectors is {\it symmetric}.

Below, the rationale underlying the choice of the similarity normalization factor mentioned above is briefly illustrated.

When computing the similarity degree of two resources $r_1$ and $r_2$, where $r_1$ and $r_2$ are annotated with the annotation vectors $av_1$ composed of $n_1$ concepts, and $av_2$ composed of $n_2$ concepts, respectively, two cases can be distinguished: either the two annotation vectors have the same cardinality, or they have different cardinalities. 

In the first case, that is $n_1$ = $n_2$, each concept in $av_1$ can be matched with one concept in $av_2$, and vice-versa. Therefore, the four options lead to the same normalization factor, and the similarity degree is computed by considering the whole semantic descriptions of both the resources.
In the second case, assuming for instance $n_1 > n_2$, there is a part of information about $av_1$ (i.e., $n_1 - n_2$ concepts) that is ignored when computing the similarity value. 

In the following, the effects of the four proposed normalization factors are illustrated when $n_1 \neq n_2$, for instance with $n_1 > n_2$. 
If the normalization factor is chosen as the maximum between $n_1$ and $n_2$, that is $n_1$, we intend to privilege richer annotations, and therefore the ``missing information'' in $av_2$ weakens the similarity between the resources.
If the normalization factor is chosen as the minimum between $n_1$ and $n_2$, that is $n_2$, we assume that a more ``compact'' annotation vector holds the essence of the resource $r_1$, and the remaining concepts are redundant.
In particular,  the choice of the normalization factor as the maximum considers the missing information (in the shorter annotation) as a deficiency. Conversely, the choice of the normalization factor as the minimum deems the additional information (in the longer annotation) as a redundancy. Therefore, the choice about the maximum or the minimum emphasize differences and commonalities between the compared annotation vectors, respectively.

Choosing the normalization factor as the arithmetic average implies a compromise between the two above cases. This case somehow considers the missing and the redundant information in annotating the compared resources in the same manner. Finally, the geometric average behaves essentially as the arithmetic average but is more sensitive to small values.

\begin{algorithm} [h!]
\renewcommand{\thealgorithm}{}
\caption{The \textit{semsim$^p$} semantic similarity function.}
\label{Semsimp-Algorithm}
\begin{algorithmic}[1]
\Function{semsim-p}{\texttt{T, av1, av2, weightingMethod, normOp}}
\If{\texttt{weightingMethod} is an extensional method}
\State \texttt{avs = loadAnnotationVectors()};
\State \texttt{T} = \texttt{weightingTaxonomy(T, weightingMethod, avs);}
\Else 
\State \texttt{T} = \texttt{weightingTaxonomy(T, weightingMethod);}
\EndIf
\State \texttt{consimMatrix[][] = empty matrix;}
\For{\texttt{i=0 to av1.size()-1}}
    \For{\texttt{j=0 to av2.size()-1}}
        \State \texttt{consimMatrix[i][j] = consim(av1[i], av2[j], T);} 
    \EndFor
\EndFor

\State \texttt{bestPairing = hungarianAlgorithm(consimMatrix);}

\State \texttt{temp = 0};

\ForAll{\texttt{bp in bestPairing}}
    \State \texttt{temp = temp + bp.value};
\EndFor
\State \texttt{normFactor = normalization(av1.size(), av2.size(), normOp);}
\State \texttt{return temp / normFactor;}
\EndFunction
\end{algorithmic}
\end{algorithm}

The algorithm for the $semsim^p$ function relies on the Hungarian algorithm for solving the {\it maximum weighted matching} problem in bipartite graphs \citep{Dulmage}, and is illustrated in the following in pseudocode.
The function takes as input a taxonomy (i.e., \texttt{T}), the annotation vectors of the two documents to be compared (i.e., \texttt{av1} and \texttt{av2}), a weighting method (i.e., \texttt{weightingMethod}) that is one out of AF, CF, TD, and IIC, and \texttt{normOp} that specifies which normalization function out of $min$, $max$, $ave$, and $gav$ must be applied.

The first step is the assignment of weights to concepts in the taxonomy (lines 2-7).
In case the weighting method is an extensional one (i.e., AF or CF), all the annotation vectors are loaded in the \texttt{avs} variable, and the \texttt{weightingTaxonomy} function is invoked with the taxonomy, the weighting method and the annotation vectors (lines 3-4).
In case the weighting method is an intensional one (i.e., TD or IIC), the annotation vectors are not needed to assign weights to concepts, and therefore the \texttt{weightingTaxonomy} function is invoked only with the taxonomy and the weighting method.
Then, the similarity values between each concept from \texttt{av1} and \texttt{av2} (i.e., \texttt{av1[i]} and \texttt{av2[j]}) are computed and stored in the \texttt{consimMatrix} (lines 8-13).
The \texttt{hungarianAlgorithm} function is invoked with the \texttt{consimMatrix} to identify the pairing between concepts from \texttt{av1} and \texttt{av2} that maximizes the sum of the similarity values (line 14). 
The sum of the similarity values between the paired concepts is computed and the value is stored in the \texttt{temp} variable (lines 15-18).
The function ends by dividing the value of the \texttt{temp} variable by the computed normalization factor, i.e, \texttt{normFactor} (lines 19-20). 

Independently of the chosen weighting method and normalization factor, in accordance with the computational complexity of the Hungarian algorithm, the \textit{semsim$^p$} method is polynomial in the size of $n$, specifically $O(n^3)$, where $n$ is the cardinality of the larger of \texttt{av1} and \texttt{av2}.

\subsection{Ontology Weighting Methods in \textit{SemSim$^p$}} \label{sect:weighting}

\noindent In this section, we illustrate two types of methods, namely  {\it extensional} and  {\it intensional}, which are introduced to compute the probability of concepts (weights) in a tree-shaped taxonomy.
Extensional methods compute concept weights by taking into account both the topology of the ISA hierarchy and the content of the annotated dataset.
Intensional methods allow concept weights to be derived on the basis of the sole topology of the ISA hierarchy. 
The former need a statistically significant number of annotated resources and provide more adherence to the current state of the reality, whereas the latter can always be applied.
These two types of methods are illustrated with the support of a running example based on the toy ontology shown in Figure \ref{fig:small-ont}, and the dataset composed of the four annotated resources, namely $r_1$, $r_2$, $r_3$, and $r_4$, shown in Table \ref{tab:simpledataset}.

\begin{table}[h]	
	\caption{Running example dataset}
	\label{tab:simpledataset}
	\vspace*{0.1in}
	\centering
	\begin{tabular}{ll} 
		\hline
		Resource & Annotation Vector \\ \hline
		$r_1$ & $av_1 = (Worker, Student)$ \\
		$r_2$ & $av_2 = (Employee)$ \\
		$r_3$ & $av_3 = (Student)$ \\
		$r_4$ & $av_4 = (Employee, Freelance)$ \\\hline
	\end{tabular}
\end{table}


\subsubsection{Extensional Methods}\label{sec:ExtensionalMethods}

The extensional methods illustrated in this section are the Concept Frequency (CF) and the Annotation Frequency (AF).

\noindent \textbf{Concept Frequency method}. The  CF method is based on the standard approach of computing the relative frequency of a concept from a taxonomy in a corpus of documents \citep{Resnik1995}. According to this approach, given a concept \textit{c}, its relative frequency (referred to as $w_{CF}(c)$ weight) is the number of occurrences of $c$ and its descendants, divided by the number of occurrences of all concepts in the set of all the annotation vectors. Formally we have:
\begin{equation} \label{eq:prob_M1}
w_{CF}(c) = \frac{n(c^+)}{N},
\end{equation}

\noindent where $c^+$ is the set formed by  $c$ and its descendants in the taxonomy, $n(c^+)$ is the total number of occurrences of the concepts in $c^+$, and $N$ is the total number of occurrences of the concepts in all the annotation vectors of the dataset.



For instance, considering the running example, in the case of the concept \textit{Worker}, the \textit{Worker$^+$} set is  \{\textit{Worker}, \textit{Employee}, \textit{Freelance}\} and $n(Worker^+$) is equal to 4. In fact, considering the annotation vectors of Table \ref{tab:simpledataset}, the concepts \textit{Worker} and \textit{Freelance} appear once, and the concept \textit{Employee} appears twice. Furthermore, the total number of occurrences of all the concepts in the set of the four annotation vectors is equal to 6. Consequently:
$w_{CF}(Worker) = \frac{4}{6} = \frac{2}{3}$.
\\









\noindent \textbf{Annotation Frequency method}. The AF method is  basically inspired by the well-known notion of \textit{inverse document frequency} (\textit{IDF}).  It is a part of the \textit{term frequency} ($TF$)-$IDF$ notion used in information retrieval to evaluate the relevance of a term in a document, extracted from a corpus of documents.

Given a concept \textit{c}, its \textit{IDF} is computed as the {\it logarithm} of the ratio of the number of documents in the collection to those containing \textit{c}:
\begin{equation} \label{eq:idf}
IDF(c) = \log_b{\frac{|AV|}{|AV_{c^+}|}}
\end{equation}
where, AV is the set of all the annotation vectors in the dataset, and $AV_{c^+}$ is the subset of AV containing the concept $c$ or at least one of its descendants.

 In our case, given a concept $c$, its relative frequency (referred to as $w_{AF}(c)$ weight) in the $AF$ method is the number of annotation vectors containing $c$ or a descendant of its, divided by the total number of annotation vectors, as follows:

\begin{equation} \label{eq:prob_M2}
w_{AF}(c) = b^{-IDF(c)} = \frac{|AV_{c^+}|}{|AV|}
\end{equation}


\noindent where, according to our approach, $b=e$.

Considering the concept \textit{Worker} in the running example, according to the $AF$ method, we have that 
$|AV_{Worker^+}|$ is equal to 3, because the concept \textit{Worker} and its descendants, namely \textit{Employee} and \textit{Freelance}, appear in 3 annotation vectors, namely $av_1$, $av_2$, and $av_4$.
Therefore, since 4 is the total number of annotated resources, the following holds:
$w_{AF}(Worker) =  \frac{3}{4}$.







\subsubsection{Intensional Methods} \label{sect:intensional}

The intensional methods illustrated in this section are the Top-Down topology-based (TD) and the Intrinsic Information Content (IIC).

\noindent \textbf{Top-Down topology-based method}. The TD method has been introduced in \cite{FMPT08} and extensively experimented in \cite{FMPT13}, where it has been referred to as \textit{probabilistic} method.
It essentially computes the probabilities of concepts in the reference ontology by adopting a {\it uniform probabilistic distribution} along the ISA hierarchy following a top-down approach. Specifically, the root of the ISA hierarchy has the probability equal to 1, and the probability of a concept $c$ (referred to as $w_{TD}(c)$) of the ontology is obtained as follows:

\begin{equation} \label{eq:M4}
w_{TD}(c) = \frac{w(parent(c))}{|siblings(c)+1|}
\end{equation}

\noindent In the running example, according to this approach, we have
$w_{TD(Worker)}$ = $\frac{1}{2}$,
 since \textit{Worker} is one of the two children of the root \textit{Person}.
\\



\noindent \textbf{Intrinsic Information Content method}. The IIC method  has been conceived to compute the information content of a concept in a taxonomy \citep{Seco2004} defined as a function of its descendants. In particular, the basic assumption is that the more descendants a concept has the less information it expresses. Therefore, leaves in the taxonomy are the most specific ones, and consequently, their information contents are maximal.

Given a taxonomy, the Intrinsic Information Content ($iic$) of a concept $c$ is given as follows:

\begin{equation} \label{eq:M4}
iic(c) = 1- \frac{log(|desc(c)| + 1) }{log(|\mathcal C|)}
\end{equation}

\noindent where $desc(c)$  is the set of the descendants of the concept $c$, and $\mathcal C$ is the set of the concepts in $Ont$.

Note that the denominator in Eq. (\ref{eq:M4}) ensures the $iic$ values are in  $[0,1]$. 
Furthermore, the information content of the root node in the taxonomy is equal to 0.\\
\noindent Now let us consider the taxonomy shown in Figure \ref{fig:small-ont}. The intrinsic information content of the concept \textit{Worker} is defined as follows:
$iic(Worker) = 1-\frac{log(2+1)}{log(5)} = 0.32$
since the descendants of \textit{Worker} and the total number of concepts in the taxonomy are 2 and 5, respectively.

\section{Case Study and Experiments} \label{sect:case-study}

\subsection{Context description} \label{sect:contect-description}
The first domain of our experimentation concerns the papers published in computer science  by the ACM and, in particular, appeared in two journals, which are the ACM Transactions on Information Systems (TOIS) and the ACM Transactions on Database Systems (TODS) journals. The total number of articles we considered in the experiment is 1,103, which corresponds to the number of papers published in these journals in the time window from  January 1990 to June 2017. 
One of the primary goal of the Digital Library of the Association for Computing Machinery (ACM-DL) is the quick discovery and the easy accessibility to the content of the articles by the reader. For this reason,  most of the papers published in the ACM journals were manually annotated by the authors according to the ACM Computing Classification System (ACM-CCS).  
The latest version of the ACM-CCS was developed in 2012. This replaces the traditional 1998 version, and is used as one of the de facto standard classification systems in the computing field.
It consists of more than 2,000 subjects covering all the fields of computer science. These subjects are not mutually exclusive and reflect the state of the art of the computing discipline, therefore they are supposed to be expanded and updated with the forthcoming subject areas in computer science. 

\subsection{Research Questions and Indicators }
\label{sec:Research_Questions}
We set up 16 different configurations of \textit{SemSim$^p$}. Each configuration is characterized by one of the previously mentioned four methods to compute the weights of the concepts (i.e., CF, AF, TD, and IIC), and  one of the four different options to compute the similarity normalization factor (i.e., $max$, $min$, $ave$, and $gav$).
Furthermore, in this analysis we considered the Dice, Jaccard, Sigmoid, WNSim similarity methods and the ones proposed respectively by Rezaei \& Fr{\"a}nti \cite{Reza2014} and by Haase et al. \cite{Hasse2004} with the purpose of comparing them with \textit{SemSim$^p$}.

In order to evaluate the performance of these methods, we identified the following research questions: 
\begin{itemize}
\item \textbf{RQ1.}  Assuming that we have a collection of papers selected by following some similarity criteria,
 which is the similarity method that allows us to assert with the highest degree of certainty that these papers were not gathered together randomly?
\item \textbf{RQ2.} 
Assuming that we have a limited number of papers, in order to identify their mutual similarity, which is the method that provides results closer to those given by human experts?
\end{itemize}
To the purpose of answering to these research questions, we defined two indicators, namely the degree of confidence, indicated also as \textbf{I1}, for \textbf{RQ1}, and the Pearson correlation, indicated also as \textbf{I2}, for \textbf{RQ2}.

\textbf{I1} is based on a statistical inference method, called \textit{null hypothesis significance testing} \citep{mertens2019new}, by which an hypothesized factor (i.e., asserted hypothesis) is tested against an hypothesis of no effect or relationship (i.e., null hypothesis) on the basis of empirical observations \citep{pernet2017}.
In our work, the \textit{hypothesis} ``the papers in a given collection are similar'' is compared to 
the \textit{null hypothesis} that ``they are not similar'' since they have been selected randomly.
Assuming that the papers  in a given collection are similar, we compare the similarity methods against the given assumption, then the best similarity method is the one that allows to assert with the highest probability that the papers in the collection are similar. Since the probability that the asserted hypothesis is true corresponds to the degree of confidence (\textbf{I1}), the higher \textbf{I1}, the better the method.





In order to answer to the second research question, we asked five experts to evaluate the mutual semantic similarity of the papers in the given sets, and we computed the Pearson correlation (\textbf{I2}) between similarity methods' results and expert judgement.

\subsection{Study settings} \label{sec:Studysettings}

As anticipated in Section \ref{sec:Introduction}, in the presence of a large dataset, it is unrealistic to  ask a group of people to pairwise check thousands of resources, producing millions of similarity scores. For this reason, 
we selected some sets of papers  among  the 1,103 TOIS and TODS papers mentioned above. In particular, we focused our attention on the seven TOIS special issues published in the aforementioned time window, i.e., from January 1990 to June 2017\footnote{In this time interval, no special issue was published on the TODS journal.}. A special issue can be considered as a {\it set} of papers selected by the editor on the basis of their topics, i.e., on the basis of the similarity of their contents, which should be determined by the research topic indicated in the call for paper.
Hence, we assumed that the mutual similarity of special issue papers has been already implicitly validated by the corresponding editors.
Therefore, we performed a first experiment validating \textit{SemSim$^p$} against the editor decision in assembling papers in each special issue. 
In particular, we expect that a special issue gathers papers with semantic cohesion significantly higher than the average of  a sufficiently great number (i.e., 100,000) of sets of papers, randomly sampled from the whole set of 1,103 papers, having the same cardinality as the special issues.

Successively, in a second experiment, we asked some experts in computer science for an assessment of the similarity of the papers contained in each of the seven special issues. In particular, for each special issue, five experts of our research institutes were asked to evaluate the mutual similarity of the papers of the special issue. This evaluation was performed at two levels: first by addressing  the ACM-CCS keywords of the papers, then also by analyzing  their abstracts, because sometimes the keywords do not represent the content of a paper in an exhaustive way.  
The overall experimentation set up was organized according to two phases: preprocessing of the ACM dataset and  experimental set up\footnote{The ACM dataset used for performing the experimentation can be accessed at the following link: \url{http://dx.doi.org/10.17632/r4vbkhgxx3.2} \cite{https://doi.org/10.17632/r4vbkhgxx3.2}}.

\subsubsection{Preprocessing of the ACM dataset} \label{DatasetPreprocessing}

This phase was devoted to acquire the ACM data and prepare them in order to be used to perform the experiments. It consists of six steps:
 
\begin{itemize}
\item Collection of the ACM papers and keywords extraction;
\item Engineering the ontology derived from ACM-CCS;
\item Semantic annotation of the ACM papers;
\item Weighting the ontology concepts;
\item Computation of similarity values;
\item Selection of samples of mutually similar papers. 
\end{itemize}

\noindent \textbf{Collection of the ACM papers and keywords extraction.} During this step, the papers of the TOIS and TODS journals in the mentioned time range were collected. Afterwards, the ACM-CCS keywords were extracted  by means of basic text processing techniques. In general authors who want to publish with the ACM are required to provide and to associate with their articles the appropriate keywords from the ACM-CCS. 
The ACM provides an accurate categorization of the papers since the keywords are checked first by the reviewers during the revision process and, successively, by the editors of the journals, before publication. 

The result of this step is a text file containing the list of papers published from January 1990 to June 2017 in the TODS and TOIS journals with the corresponding ACM-CCS keywords. Each item of the list includes a numeric identifier, the year of publication, the title of the paper, the author(s), and the ACM-CCS keywords.

\noindent \textbf{Engineering the ontology derived from ACM-CCS.}
\label{ACM_ontology_engineering}
The aim of this step was to build an ontology \citep{denicola2016, denicola2009, ESPINOZAARIAS2021100655} in an appropriate format to compute semantic similarity values by means of the \textit{SemSim$^p$} algorithm.
In the ACM, the CCS is a
Directed Acyclic Graph (DAG) \citep{Schri2003}, where each node is labeled with a subject area of computer science, and arcs denote a hierarchical relationship.
As mentioned in Section \ref{sect:semsimp}, the \textit{SemSim$^p$} method requires a tree-shaped  ontology to compute the similarity values.
Therefore, to the purpose of the experiment, we built a tree-shaped taxonomy on the basis of the ACM-CCS as follows. For each path in the ACM-CCS  ending in a root (i.e., a node without parents), we added a concept in the ontology, which is identified by the concatenation of the labels of the ACM-CCS nodes along the considered path. 
Accordingly, if a node in the ACM-CCS has two parents, it has two paths ending in some roots and leads to the generation of two concepts in the ontology, in order to remove the multiple inheritance. Then, each of the new concepts is linked to only one parent according to the specialization relationship.
For instance, in the ACM-CCS, both the nodes {\small \texttt{``Cross-computing  tools  and  techniques''}}\footnote{Note that, for the sake of clarity, ACM-CCS keywords are in quoted \texttt{Courier}, whereas the  concepts in the developed ontology are in quoted \textit{italic}.},  and {\small \texttt{``Dependable  and  fault-tolerant  systems  and  networks''}} have  {\small \texttt{``Reliability''}} as a child. In turn, these two parents  have {\small \texttt{``General and Reference''}}, and {\small \texttt{``Computer systems organization''}} as parents, respectively, which are roots. Therefore, the following concepts were added to the ontology: {\small \textit{``Reliability\_Cross-computing tools and techniques\_General and Reference\_owl:Thing''}}, and {\small \textit{``Reliability\_Dependable and fault-tolerant systems and networks\_Computer systems organization\_owl:Thing''}}, where {\small \textit{``Thing''}} is the unique root of the   ontology we constructed. Furthermore, the former concept is a specialization of {\small \textit{``Cross-computing tools and techniques\_General and Reference\_owl:Thing''}} because {\small \texttt{``Reliability''}} is a child of {\small \texttt{``Cross-computing tools and techniques''}} in the ACM-CCS.

A fragment of the ACM-CCS 2012 is shown in Figure \ref{fig:acm_excerpt} in Annex B, in particular, part of the sub-tree having {\small \texttt{``Information systems''}} as a root.
Therefore, the {\small \texttt{``Information systems''}} node has five children ({\small \texttt{``Data management systems''}}, {\small \texttt{``Information systems applications''}}, {\small \texttt{``Information storage systems''}}, {\small \texttt{``World Wide Web''}}, and {\small \texttt{``Information retrieval''}}). 
Note that the relationship between a child and a parent node is a ISA relationship because we assumed the corresponding concepts as \textit{competences} on the topics related to the ACM-CCS keywords.
For instance, an expert in {\small \texttt{``Database management system engines''}} has also competences in {\small \texttt{``Data management systems''}} (which is the computer science area labeling its parent node) and has also competences in the areas labeling all its ancestors along the ISA hierarchy, therefore in this case also in	{\small \texttt{``Information systems''}}, which is one of the root nodes of the ACM-CCS.

In our experiment, the online categorization revised by the ACM was used as published in June 2017.  
The result of this step is an ontology on computer science concepts, organized as a tree and available in the OWL/RDF format\footnote{\url{https://www.w3.org/TR/owl-ref/}}. It consists of 2,113 concepts and 2,114 ISA relationships.
The ontology was automatically built from an html version of the ACM-CCS keywords.

\noindent \textbf{Semantic annotation of the ACM papers.} 
This step was devoted to annotate all the selected papers with the concepts of the ontology that was built in the previous step, and to associate with each paper the corresponding annotation vector. 
Consider, for instance, one of the papers of the Special Issue 3 on {\it User Interface Software and Technology} and, in particular, the paper labeled Title 1, as shown in Table \ref{tab:papers_of_special_issue_III}. This paper has been categorized by the authors by specifying the following keywords (referred to as  \textit{Categories and Subject Descriptors} in the ACM):\\

\begin{small}
\noindent {\footnotesize \texttt{C.2.4 [Computer-Communication Networks]: Distributed Systems - distributed \indent applications, distributed databases;}}\\ 
\noindent {\footnotesize \texttt{D.2.2 [Software Engineering]: Tools and Techniques - user interfaces;}}\\
\noindent {\footnotesize \texttt{D.2.6 [Software Engineering]: Programming Environments - user  interactive;}}\\ 
\noindent {\footnotesize \texttt{D.3.3 [Programming Languages]: Language Constructs - input/output;}}\\ 
\noindent {\footnotesize \texttt{H.1.2 [Models and Principles]: User/Machine Systems - human factors;}}\\ 
\noindent {\footnotesize \texttt{H.4.1 [Information Systems Applications]: Office Automation;}}\\ 
\noindent {\footnotesize \texttt{1.7.1 [Text Processing]: Text Editing.}}\\
\end{small}

 \begin{table*}[h] 
	\centering 
	\caption{Papers of the Special Issue 3.}
	\label{tab:papers_of_special_issue_III} 
	\vspace*{0.1in}
	\scriptsize 
	\begin{tabular}{| l | c | c |} 
		\hline 
		\textbf{Title} & \textbf{Short title} 
		\tabularnewline 
		\hline 
		\multirow{2}{250pt}{A High-Level and Flexible Framework for Implementing Multiuser User Interfaces.}    & \multirow{2}{*}{\textbf{Title 1}}  \tabularnewline 
		& \tabularnewline
		\hline 
		\multirow{2}{250pt}{A Model for Input and Output for Multilingual Text in a Windowing Environment.} & \multirow{2}{*}{\textbf{Title 2}}   \tabularnewline 
		& \tabularnewline
		\hline 
		\multirow{2}{250pt}{EmbeddedButtons: Supporting Buttons in Documents.} & \multirow{2}{*}{\textbf{Title 3}}  \tabularnewline 
		& \tabularnewline
		\hline 
		\multirow{2}{250pt}{Lessons Learned from SUIT, the Simple User Interface Toolkit.} & \multirow{2}{*}{\textbf{Title 4}}  \tabularnewline 
		& \tabularnewline
		\hline 
		\multirow{2}{250pt}{A General Framework for Bidirectional Translation between Abstract and Pictorial Data.}   & \multirow{2}{*}{\textbf{Title 5}}   \tabularnewline 
		& \tabularnewline
		\hline 
		
	\end{tabular} 
	\centering 
\end{table*}

Since this paper was published in 1992, these keywords have been specified according to  previous versions of the  ACM-CCS 2012. Successively, after the release of the 2012 version, paper categorizations have been revised by the ACM and, for each paper, the list of updated keywords can be found online in the {\it index term} menu, where classification numbers (such as {\small \texttt{C.2.4}} or {\small \texttt{D.2.2}} above) have been omitted. For instance, in the previous example, the category {\small \texttt{H.4.1 [Information Systems Applications]: Office Automation;}} has been replaced with the following one: {\small \texttt{[Information Systems]: Information Systems Applications, Enterprise information systems - enterprise applications}}.
Therefore, in our experiment, the ontology concept corresponding to the above ACM-CCS path has the following identifier: {\small \textit{``Enterprise applications\_Enterprise information systems\_Information systems applications\_Information systems\_owl:Thing''}}.

\noindent \textbf{Weighting the ontology concepts.}
As mentioned in the previous sections, in the \textit{SemSim$^p$} approach, each node has to be associated with a weight. To this aim, this step was devoted to evaluate the weights of the ACM concepts according to the CF, AF, TD, and IIC methods presented in Section \ref{sect:weighting}. 


In Figure \ref{fig:ACMTax}, the ontology derived from the ACM-CCS is drawn as a graph, where nodes and arcs represent concepts and  ISA relationships, respectively. In particular,
the size of each node is proportional to the associated weight computed according to the extensional (CF and AF) and the intensional (TD and IIC) methods. The {\small\textit{owl:Thing}}\footnote{\textit{owl:} stands for the OWL namespace, i.e., \url{http://www.w3.org/2002/07/owl\#}.} node, i.e., denoting the most general concept, is white and the other ones are green. 
In Figure \ref{fig:ACMTaxZoom}, the fragment of the ontology related to the information systems subject area, corresponding to the bottom right sides of the graphs of Figure \ref{fig:ACMTax}, is detailed.
Note that, due to our focus on the TOIS and TODS journals, i.e.,  the information systems area, the 
sizes of the nodes corresponding to the extensional methods are more heterogeneous with respect to 
those corresponding to the intensional ones, whose weights depend only on the structure of the taxonomy. 
Figure \ref{fig:ACMTax} and Figure \ref{fig:ACMTaxZoom} graphically show how the sizes of the nodes, corresponding to the weights of the concepts, differ depending on the different weighting methods. Note that, in general, different weights lead to  different similarity values of the \textit{SemSim$^p$} method, according to the Eq. (\ref{eq:consim}), on which it is based.
Both the figures were realized by using the  Gephi\footnote{\url{https://gephi.org}} software \citep{ICWSM09154}.


\begin{figure}[h]
	\includegraphics[width=\linewidth]{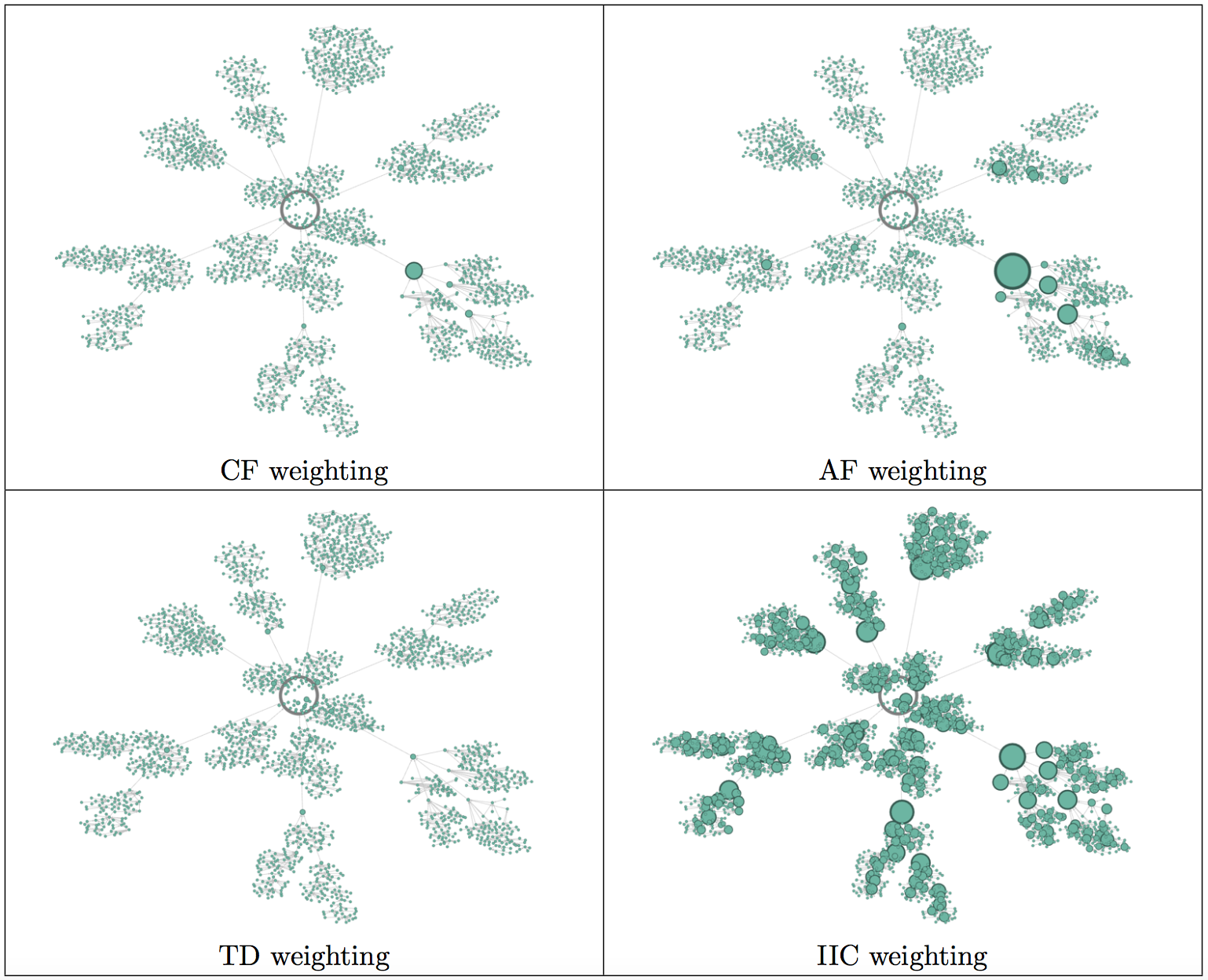}
	\caption{Ontology derived from the ACM-CCS with the node sizes proportional to the weights computed by means of the extensional CF, AF, and intensional TD, IIC weighting methods.}
	\label{fig:ACMTax}
\end{figure}

\begin{figure}[h]
	\includegraphics[width=\linewidth]{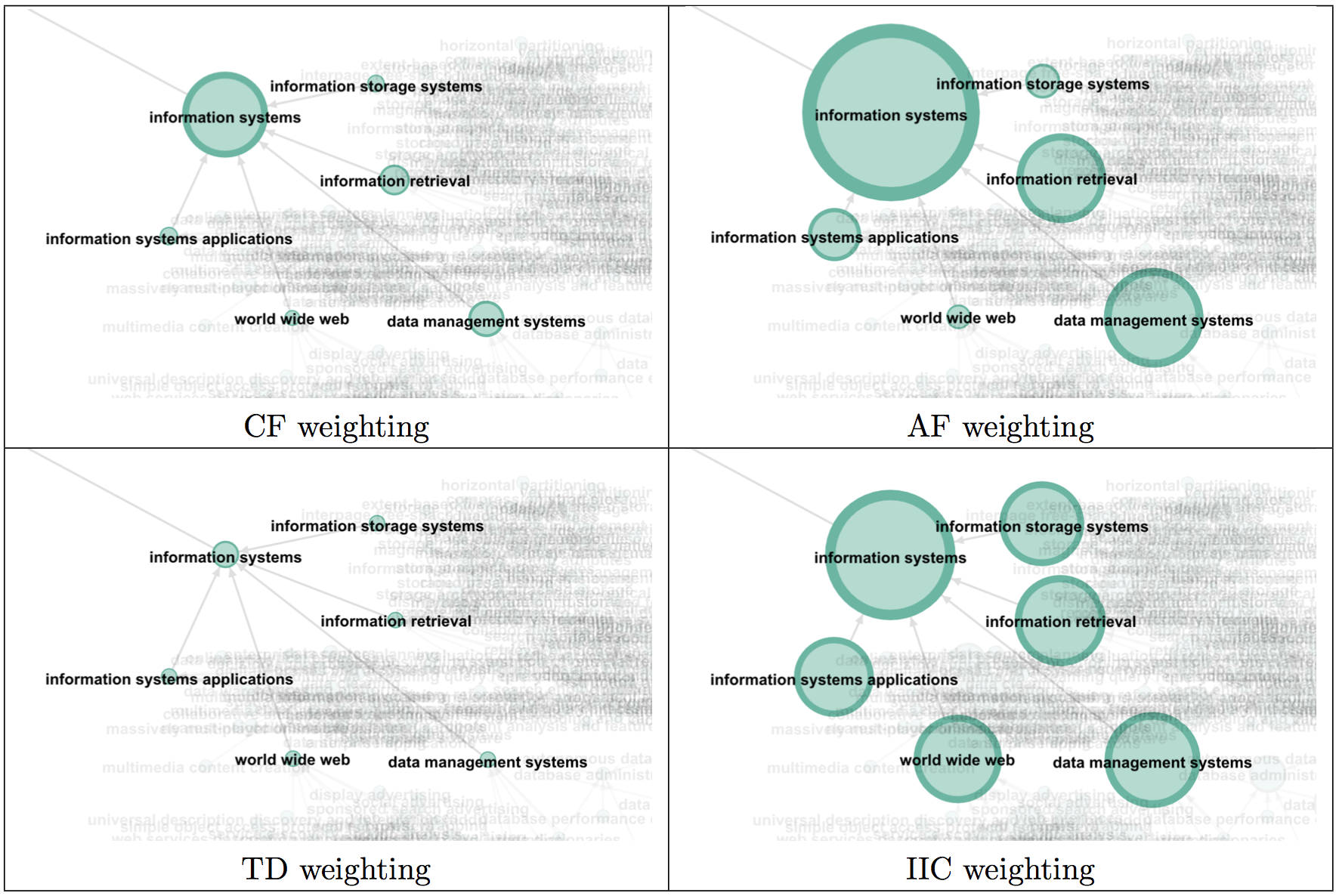}
	\caption{An excerpt from the ontology derived from ACM-CCS related to the \textit{information system} sub-tree, with the node sizes proportional to the weights computed by means of the extensional CF, AF, and intensional TD, IIC weighting methods.}
	\label{fig:ACMTaxZoom}
\end{figure}

\noindent {\bf Computation of similarity values.} \label{SimilarityComputation}
This phase was devoted to compute the similarity matrices containing the semantic similarity values of all possible pairs of the collected papers according to the 16 \textit{SemSim$^p$} configurations, as well as the other selected methods.
As the number of collected papers is 1,103, each resulting similarity matrix has more than 1,000,000 values.

\noindent \textbf{Selection of samples of mutually similar papers.}
This step was devoted to select samples of papers that are mutually similar. As mentioned before, we assumed that special issues were assembled together by editors following some similarity criteria, and we addressed all the special issues from January 1990 to June 2017. The titles of the special issues are presented in Table \ref{tab:special_issue_data_I}, whereas in Table \ref{tab:special_issue_data_II}, the number of papers of each special issue and the number of ACM keywords of each paper are shown. As illustrated before, for the sake of brevity, only the titles of the papers included in the Special Issue 3 are shown in Table \ref{tab:papers_of_special_issue_III}.\\


\begin{table*}[h] 
	\centering 
		\caption{Titles of the selected TOIS special issues.} 
	\label{tab:special_issue_data_I}
	\vspace*{0.1in} 
	\scriptsize 
	\begin{tabular}{| l | c | c |} 
		\hline 
		\textbf{Title} & \textbf{Short title} 
		\tabularnewline 
		\hline 
		\multirow{2}{250pt}{Special Issue on Computer-Human Interaction.\\Volume 9 Issue 2, April 1991}    & \multirow{2}{*}{\textbf{Sp.Iss.1}}  \tabularnewline 
		& \tabularnewline
		\hline 
		\multirow{2}{250pt}{Special Issue on Research and Development in Information Retrieval.\\Volume 9 Issue 3, July 1991} & \multirow{2}{*}{\textbf{Sp.Iss.2}}   \tabularnewline 
		& \tabularnewline
		\hline 
		\multirow{2}{250pt}{Special Issue on User Interface Software and Technology.\\Volume 10 Issue 4, Oct. 1992} & \multirow{2}{*}{\textbf{Sp.Iss.3}}  \tabularnewline 
		& \tabularnewline
		\hline 
		\multirow{2}{250pt}{Special Issue on Social Science Perspectives on IS.\\Volume 12 Issue 2, April 1994} & \multirow{2}{*}{\textbf{Sp.Iss.4}}  \tabularnewline 
		& \tabularnewline
		\hline 
		\multirow{2}{250pt}{Special Issue on Video Information Retrieval.\\Volume 13 Issue 4, Oct. 1995}   & \multirow{2}{*}{\textbf{Sp.Iss.5}}   \tabularnewline 
		& \tabularnewline
		\hline 
		\multirow{2}{250pt}{Special Issue on Contextual Search and Recommendation.\\Volume 33 Issue 1, March 2015} & \multirow{2}{*}{\textbf{Sp.Iss.6}}   \tabularnewline 
		& \tabularnewline
		\hline 
		\multirow{2}{250pt}{Special Issue on Trust and Veracity of Information in Social Media.\\Volume 34 Issue 3, May 2016} & \multirow{2}{*}{\textbf{Sp.Iss.7}}  \tabularnewline 
		& \tabularnewline
		\hline 
	\end{tabular} 
	\centering 
\end{table*}

\begin{table*}[h] 
	\centering 
		\caption{Number of papers (\# papers) of the selected TOIS special issues and number of ACM keywords (\# keywords) of each paper. NA means that the paper 5 is not available because the special issue contains 4 papers.} 
	\label{tab:special_issue_data_II} 
	\vspace*{0.1in} 
	\scriptsize 
	\begin{tabular}{| c  | c | c | c | c | c | c | c |} 
		\hline 
		&  \rotatebox{90}{\textbf{Sp.Iss.1~}} & \rotatebox{90}{\textbf{Sp.Iss.2}} & 
		\rotatebox{90}{\textbf{Sp.Iss.3}} & \rotatebox{90}{\textbf{Sp.Iss.4}} &
		\rotatebox{90}{\textbf{Sp.Iss.5}} & \rotatebox{90}{\textbf{Sp.Iss.6}} &
		\rotatebox{90}{\textbf{Sp.Iss.7}} 
		\tabularnewline 
		\hline 
		
		\# papers   & 4 & 4 & 5 & 4 & 4 & 5 & 5  \tabularnewline 
		\hline 
		\hline 
		\# keywords of paper 1    & 4 & 4 & 10 & 1 & 7 & 1 & 1  \tabularnewline 
		\hline 
		\# keywords of paper 2    & 2 & 2 & 7 & 2 & 4 & 3 & 1  \tabularnewline 
		\hline 
		\# keywords of paper 3    & 7 & 8 & 6 & 7 & 6 & 4 & 2  \tabularnewline 
		\hline 
		\# keywords of paper 4    & 5 & 6 & 9 & 3 & 8 & 1 & 4  \tabularnewline 
		\hline 
		\# keywords of paper 5    & NA & NA & 3 & NA & NA & 1 & 2  \tabularnewline 
		\hline 
	\end{tabular} 
	\centering 
\end{table*} 


\subsubsection{Experimental set up} \label{SemanticCohesion}
This phase consists of two steps. The former concerns the set up of the semantic cohesion experiment, and the latter the expert judgement experiment.

\noindent
{\bf Setting up the semantic cohesion experiment.}
This step concerns the statistical analysis in order to quantify the semantic cohesion of each special issue 
(i.e., the average of the similarity values of all pairs of papers in the issue).
 As shown in Table \ref{tab:special_issue_data_II}, each selected special issue consists of  either 4 or 5 papers. Accordingly, two sets of 100,000 samples of 4 and 5 papers were randomly selected, respectively.  

For each special issue, the semantic cohesion was computed by applying the \textit{SemSim$^p$} configurations, as well as the other selected methods. Then, the same computations were performed for each of the randomly selected samples of 4 and 5 papers, and the average similarity values were used to build different frequency distributions.
Finally, we performed the hypothesis testing for the  null and asserted hypotheses (see Section \ref{sec:Research_Questions}).
 In particular, the {\it t-value} and the corresponding  degree of confidence of the asserted hypothesis were computed for each of the considered similarity methods and for each special issue.
The {\it t-value} ($t$) was computed as follows: 

\begin{equation}
t = \displaystyle \frac{sc_{i}-\overline{sc_R}}{\sigma}
\end{equation}

 \noindent where $sc_{i}$ is the semantic cohesion (i.e., the average similarity value) of the special issue $i$, $\overline{sc_R}$ is the average of the sample semantic cohesion values (i.e. the average of the sample semantic similarity averages) and $\sigma$ is the standard deviation of the sample semantic cohesion values.
Note that, considering the population of semantic similarity values, $\sigma$ is equal to the population standard deviation divided by the square root of the sample size \citep{bluman2009elementary}.

On the basis of the t-test, we are able to identify  
the best result, i.e., the similarity method that maximizes the degree of confidence of the asserted hypothesis (\textbf{I1}) and, hence, the probability that the assertion that a special issue gathers similar papers is true.

In order to better present the rationale underlying the first experiment, Figure \ref{fig:SpIss_6_semantic-cohesion} shows the semantic cohesion distribution for the 100,000 randomly selected samples with cardinality 5, as well as the semantic cohesion of the Special Issue 6 ($sc_6$) computed by using \textit{SemSim$^p$} configured with the annotation frequency as weighting method, and the geometric average as similarity normalization factor. 
The average value of semantic cohesion for the resulting distribution is labelled by $\overline {sc}_R$, whereas $\sigma$ labels its standard deviation. It is worth noting that $sc_6$ is more than $4\sigma$ times higher than $\overline {sc}_R$. This leads to $t=5.2621$.  According to the t-test method defined in the literature and  the value of the {\it Student's t distribution} computed by using the cumulative distribution function \citep{bluman2009elementary}, the degree of confidence of the hypothesis that the Special Issue 6 is a collection of similar papers is 99.84 \% (see Table \ref{tab:results_Spiss_ALL}). The same analysis was performed for each of the semantic similarity methods considered in this paper and for each of the seven special issues.\\ 

\begin{figure}[h]
	\centering
	\includegraphics[width=\linewidth]{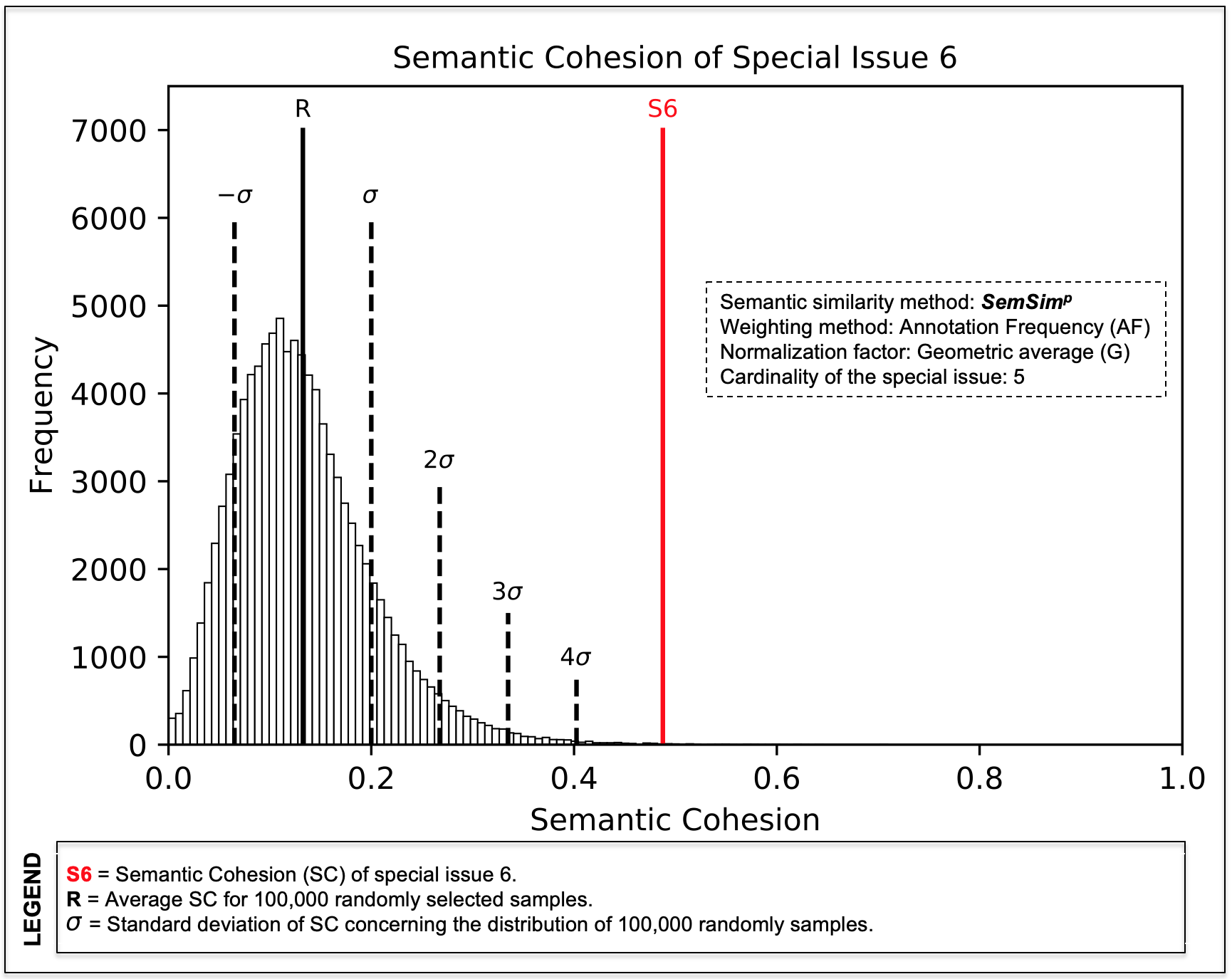}
	\caption{Statistical analysis of semantic cohesion for the Special Issue 6 computed by using \textit{SemSim$^p$} configured with the annotation frequency as weighting method, and the geometric average as similarity normalization factor.}
	\label{fig:SpIss_6_semantic-cohesion}
\end{figure}

\noindent
{\bf Setting up the expert judgement experiment.}
This step was devoted to setting up the experiment in order to determine the semantic similarity method that provides similarity values that best fit with those provided by expert judgement. To this purpose, as mentioned before, we asked five experts of our research institutes, having at least 18 years of experience in the database and information systems fields, to evaluate the mutual similarity values for all the papers included in each special issue. 
Each special issue was evaluated by three experts. The workload for each expert was assigned having in mind an objectivity-quality of result trade-off. To this aim, we avoided to overwhelm experts with heavy activities that would have diminished the quality of the evaluation.
 %
This analysis was performed at two levels: first by focusing on the ACM-CCS keywords of the papers, then by also analyzing their abstracts.
Afterwards, the Pearson correlation values (\textbf{I2}) between the
similarity values obtained from the expert-based experiment focusing on keywords and the automatically-computed ones were evaluated. The best result, in this case, is the similarity method that maximizes the Pearson correlation. 
Indeed, \textbf{I2} was evaluated by considering the experiment based on keywords because abstracts are not used by the semantic similarity methods addressed in this work. The expert-based experiment also focusing on abstracts was used in Section \ref{sect:discussion} in order to discuss the annotation quality issue.

\subsubsection{Software}

In order to support the phases presented above, we developed different modules of software by using two programming languages. In particular, the components dealing with the pre-processing of the ACM and American Physical Society (APS) datasets, which were needed for setting up the two experiments, the statistical analysis, and the component about the data visualization were implemented by using the Python programming 
 language\footnote{The code for the statistical analysis, the ACM and APS datasets, and the used ontologies are available at:
 \url{http://dx.doi.org/10.17632/pv3hvmbppk.1} \cite{https://doi.org/10.17632/pv3hvmbppk.1}}. 
 The module for the automatic generation of the ontology starting from the ACM-CCS, as well as the \textit{SemSim$^p$} method were implemented by using the Java programming language\footnote{The implementation of the \textit{SemSim$^p$} method is part of the \textit{SemanticRelatednessInRDFGraph} software project available at:\\
\url{https://github.com/ftaglino/SemanticRelatednessInRDFGraph}}.

\subsubsection{Selected similarity methods for comparison \label{sec:SelectedSimilarityMethodsForComparison}}

As mentioned, in this work, we address the similarity of the ACM papers represented by means of sets of concepts (features), i.e., their annotation vectors. In the experimentation, we selected the well-known Dice \citep{Dice1945}, and Jaccard \citep{Jaccard1912} measures and, furthermore, Sigmoid \citep{LIKAVEC2019}, WNSim \citep{Shajalal2019}, and the methods of Rezaei \& Fr{\"a}nti \citep{Reza2014}, and of Haase et al. \citep{Hasse2004}. In particular, the first three are representatives of the so-called {\it set-theoretic methods}, since they are based on set-theoretic operations applied to the 
sets of features (the paper annotation vectors), whereas the remaining three are representatives of the {\it taxonomy-based methods}, since they also rely on a taxonomy. 
These measures are formally provided in Tables \ref{tab:SetTheoreticMethods} and \ref{tab:TaxonomyBasedMethods}.  

Note that, as anticipated in Section \ref{sec:SimSetsOfConcepts}, WNSim and the method proposed by Haase et al. are asymmetric, i.e., the order used to compare two annotation vectors is relevant.
However, since in our experimentation the order of comparison is not significant,
 we computed the average of the similarity between a pair and the corresponding symmetric one. 
In Tables \ref{tab:ave_sim_values} and \ref{tab:compared_results_Spiss_ALL} and Figures \ref{fig:assessment} and \ref{fig:SpIss_7DJ}, these measures are indicated, respectively, as WNSim\textsubscript{sym} and Haase et al.\textsubscript{sym}, where \textsubscript{sym} stands for symmetric.

We did not include similarity methods based on machine learning in the comparison because most of them require access to the content of the digital resources (e.g., the text of the articles), which can not be always available. Conversely, we considered the methods that only require resources to be semantically annotated.


\subsection{Comparison results} \label{Sect:studyresults}


The results obtained from the two experiments are presented in Tables \ref{tab:ave_sim_values}, 
 \ref{tab:results_Spiss_ALL} and \ref{tab:compared_results_Spiss_ALL}. 
Note that $semsim_{AF,max}$ corresponds to the original $semsim$ function presented in \citep{FMPT13}. 
Table \ref{tab:ave_sim_values}  includes the semantic cohesion  for each special issue, computed by applying the \textit{SemSim$^p$} configurations, and the other selected methods,
and according to the expert judgement.
 {\it EJ\textsubscript{key}} stands for the expert judgement performed by only considering the ACM-CCS keywords, whereas {\it EJ\textsubscript{abs}}
considers both the keywords and the abstracts of the papers. 
Table \ref{tab:results_Spiss_ALL} illustrates, for each special issue, the degree of confidence (\textbf{I1}) and the Pearson correlation values obtained by focusing on keywords (\textbf{I2})
for all the \textit{SemSim$^p$} configurations.
For each method, the average values of all the special issues for both the indicators are shown in the ``Average'' column. 
Analogously, Table \ref{tab:compared_results_Spiss_ALL} shows the data
corresponding to the other selected methods\footnote{Given the Pearson correlation  $P= \frac{ \sum_i (s_i-\bar{s}) \cdot \sum_i (e_i-\bar{e})} { \sum_i (s_i-\bar{s})^2 \cdot \sum_i (e_i-\bar{e})^2 } $, where $s_i$ is the similarity between pairs of papers and $e_i$ is the similarity assigned by experts,  in the case of the Special Issue 7, for Dice, Jaccard, and Sigmoid, $P$ cannot be computed because $s_i = 0$ for all the pairs. In fact, the set intersection between the annotation vectors of any two papers in this special issue is empty.},  and $semsim_{AF,gav}$, which is the best performing \textit{SemSim$^p$} configuration.
In the last two tables, the best results are highlighted in bold.

For each considered method, the scatter plot in Figure \ref{fig:assessment} shows the average degree of confidence and the average Pearson correlation both computed over all the special issues. 
In the legend, the associated labels in the plot represent the similarity normalization factors (see Section \ref{sect:semsim-algo}), where
 X stands for maximum, N for minimum, A for arithmetic average, and G for geometric average, and  the weighting methods are associated with different markers (see Section \ref{sect:weighting}). 
Furthermore, $W$, $S$, $D$, $J$, $H$, $R$  standing for the WNSim\textsubscript{sym}, Sigmoid, Dice, Jaccard, Haase et al.\textsubscript{sym} and Rezaei \& Fr{\"a}nti methods, respectively, are included.

\begin{table*}[h] 
\centering 
\caption{Semantic cohesion of the selected special issues computed by similarity method and according to expert judgement. \textit{EJ\textsubscript{key}}  stands for the expert judgement performed by only considering the ACM-CCS keywords, whereas {\it EJ\textsubscript{abs}} considers both the keywords and the abstracts of the papers..} 
\label{tab:ave_sim_values} 
\vspace*{0.1in} 
\scriptsize 
\begin{tabular}{| c  | c | c | c | c | c | c | c |} 
\hline 
\textbf{Similarity method} &  \rotatebox{90}{\textbf{Sp.Iss.1~}} & \rotatebox{90}{\textbf{Sp.Iss.2}} & 
  \rotatebox{90}{\textbf{Sp.Iss.3}} & \rotatebox{90}{\textbf{Sp.Iss.4}} &
   \rotatebox{90}{\textbf{Sp.Iss.5}} & \rotatebox{90}{\textbf{Sp.Iss.6}} &
   \rotatebox{90}{\textbf{Sp.Iss.7}} 
 \tabularnewline 
\hline 
\hline 

$semsim_{AF,max}$   & 0.21 & 0.18 & 0.28 & 0.26 & 0.47 & 0.37 & 0.05  \tabularnewline 
\hline 
$semsim_{CF,max}$    & 0.22 & 0.21 & 0.28 & 0.25 & 0.48 & 0.39 & 0.06  \tabularnewline 
\hline 
$semsim_{TD,max}$    & 0.22 & 0.27 & 0.28 & 0.23 & 0.48 & 0.44 & 0.09  \tabularnewline 
\hline 
$semsim_{IIC,max}$    & 0.22 & 0.25 & 0.28 & 0.23 & 0.47 & 0.42 & 0.08  \tabularnewline 
\hline 
\hline 

$semsim_{AF,min}$    & 0.42 & 0.38 & 0.49 & 0.73 & 0.68 & 0.71 & 0.09  \tabularnewline 
\hline 
$semsim_{CF,min}$    & 0.43 & 0.45 & 0.50 & 0.71 & 0.69 & 0.76 & 0.12  \tabularnewline 
\hline 
$semsim_{TD,min}$    & 0.43 & 0.57 & 0.50 & 0.66 & 0.69 & 0.83 & 0.16  \tabularnewline 
\hline 
$semsim_{IIC,min}$    & 0.43 & 0.53 & 0.48 & 0.67 & 0.68 & 0.80 & 0.15  \tabularnewline
\hline 
\hline 

$semsim_{AF,ave}$    & 0.28 & 0.24 & 0.35 & 0.37 & 0.55 & 0.44 & 0.06  \tabularnewline 
\hline 
$semsim_{CF,ave}$    & 0.29 & 0.28 & 0.35 & 0.36 & 0.56 & 0.48 & 0.08  \tabularnewline 
\hline 
$semsim_{TD,ave}$    & 0.29 & 0.35 & 0.35 & 0.33 & 0.56 & 0.53 & 0.11  \tabularnewline 
\hline 
$semsim_{IIC,ave}$    & 0.29 & 0.33 & 0.34 & 0.33 & 0.55 & 0.51 & 0.10  \tabularnewline 
\hline 
\hline 

$semsim_{AF,gav}$    & 0.29 & 0.26 & 0.37 & 0.42 & 0.56 & 0.49 & 0.06  \tabularnewline 
\hline 
$semsim_{CF,gav}$    & 0.31 & 0.31 & 0.37 & 0.41 & 0.57 & 0.52 & 0.09  \tabularnewline 
\hline 
$semsim_{TD,gav}$    & 0.31 & 0.38 & 0.37 & 0.38 & 0.57 & 0.57 & 0.12  \tabularnewline 
\hline 
$semsim_{IIC,gav}$    & 0.30 & 0.36 & 0.36 & 0.39 & 0.57 & 0.55 & 0.11  \tabularnewline 
\hline 
\hline 
Dice   & 0.13 & 0.11 & 0.18 & 0.19 & 0.43 & 0.23 & 0  \tabularnewline 
\hline 
Jaccard   & 0.08 & 0.06 & 0.10 & 0.13 & 0.28 & 0.18 & 0  \tabularnewline 
\hline 
Sigmoid   & 0.03 & 0.03 &	0.04 & 0.06 & 0.11 & 0.08 & 0   \tabularnewline 
\hline
\hline 
WNSim\textsubscript{sym}    & 0.16 & 0.12 & 0.16 & 0.41 & 0.53 & 0.24 & 0  \tabularnewline 
\hline 
Rezaei \& Fr{\"a}nti  & 0.42 & 0.52 & 0.47 & 0.49 & 0.62 & 0.70 & 0.32  \tabularnewline 
\hline 
Haase et al.\textsubscript{sym}   & 0.41 & 0.51 & 0.45 & 0.54 & 0.64 & 0.75 & 0.25  \tabularnewline 
\hline
\hline 

{\it EJ\textsubscript{key}}  & 0.33 & 0.38 & 0.38 & 0.27 & 0.54 & 0.41 & 0.18  \tabularnewline 
\hline 
{\it EJ\textsubscript{abs}}  & 0.23 & 0.28 & 0.41 & 0.39 & 0.58 & 0.39 & 0.39  \tabularnewline 
\hline

\end{tabular} 
\centering 
\end{table*}

\begin{sidewaystable}

\caption{Degree of confidence \textbf{I1} and Pearson correlation \textbf{I2} values for each special issue obtained by using all the \textit{SemSim$^p$} configurations.
The table also includes the average values for both the indicators in the ``Average'' column.} \label{tab:results_Spiss_ALL}
\tiny
\begin{tabular}{| c  | c | c || c | c | c | c | c | c | c | c | c | c | c | c | c | c | c | c | c | c | c | c | c | c | c | c | c | c | c | c | c | c |} 
\hline 
 &  \multicolumn{2}{c|}{Average} &  \multicolumn{2}{c|}{Sp.Iss.1} & \multicolumn{2}{c|}{Sp.Iss.2} & \multicolumn{2}{c|}{Sp.Iss.3} & \multicolumn{2}{c|}{Sp.Iss.4} & \multicolumn{2}{c|}{Sp.Iss.5} & \multicolumn{2}{c|}{Sp.Iss.6} & \multicolumn{2}{c|}{Sp.Iss.7}  \tabularnewline
\hline 
\textbf{Similarity method} &  {\textbf{I1}} & {\textbf{I2}} & 
  {\textbf{I1}} & {\textbf{I2}} &
   {\textbf{I1}} & {\textbf{I2}} &
   {\textbf{I1}} & {\textbf{I2}} &
   {\textbf{I1}} & {\textbf{I2}} &
   {\textbf{I1}} & {\textbf{I2}} &
   {\textbf{I1}} & {\textbf{I2}} &
   {\textbf{I1}} & {\textbf{I2}} 
 \tabularnewline 
\hline 
\hline 

$semsim_{AF,max}$  & 84.95\% & 0.60 & 92.8\% & 0.8 & 	87.28\% & \textbf{0.54} & 	99.13\% & 0.42 & 	96.64\% & 0.15 & 	\textbf{99.78\%} & 0.42 & 	99.81\% & 0.93 & 	\textbf{19.23\%} & \textbf{0.92}  \tabularnewline 
\hline 
$semsim_{CF,max}$  & 82.18\% & 0.59 & 86.74\% & 0.78 & 	84.78\% & 0.43 & 	97.55\% & 0.46 & 	91.82\% & 0.21 & 	99.6\% & 0.4 & 	99.66\% & 0.94 & 	15.1\% & \textbf{0.92}  \tabularnewline 
\hline 
$semsim_{TD,max}$  & 75.38\% & 0.56 &  70.01\% & 0.71 & 	 83.11\% & 0.25 & 	 90.33\% & 0.51 & 	 72.5\% & 0.32 & 	 98.84\% & 0.39 & 	 99.24\% & 0.93 & 	 13.62\% & 0.78  \tabularnewline
\hline 
$semsim_{IIC,max}$  & 77.45\% & 0.58 &75.07\% & 0.73 & 	83.45\% & 0.31 & 	92.08\% & 0.55 & 	78.77\% & 0.3 & 	99.1\% & 0.42 & 	99.35\% & 0.94 & 	14.36\% & 0.82  \tabularnewline
\hline 
\hline 

$semsim_{AF,min}$   & 85.17\% & 0.61 & 92.6\% & \textbf{0.98} & 	89.85\% & 0.31 & 	98.33\% & 0.64 & 	\textbf{99.38\%} & 0.67 & 	99.15\% & 0.51 & 	99.81\% & 0.22 & 	17.04\% & \textbf{0.92}  \tabularnewline
\hline 
$semsim_{CF,min}$  & 82.52\% & 0.59 & 85.69\% & \textbf{0.98} & 	88.17\% & 0.16 & 	95.26\% & 0.63 & 	98.58\% & 0.7 & 	98.32\% & 0.5 & 	99.65\% & 0.25 & 	11.96\% & 0.91  \tabularnewline
\hline 
$semsim_{TD,min}$  & 76.03\% & 0.56 &  65.77\% & 0.96 & 	 87.09\% & -0.05 & 	 82.14\% & 0.67 & 	 93.73\% & \textbf{0.73} & 	 94.86\% & 0.49 & 	 99.09\% & 0.3 & 	 9.5\% & 0.78 \tabularnewline
\hline 
$semsim_{IIC,min}$   & 77.87\% & 0.58 &  71.64\% & 0.97 & 	87.17\% & 0.02 & 	85.06\% & 0.71 & 	95.63\% & 0.72 & 	96.01\% & \textbf{0.52} & 	99.26\% & 0.28 & 	10.31\% & 0.82 \tabularnewline
\hline 
\hline 

$semsim_{AF,ave}$   & \textbf{85.80\%} & 0.66 & \textbf{94.31\%} & 0.9 & 	90.0\% & 0.53 & 	\textbf{99.22\%} & 0.6 & 	98.35\% & 0.3 & 	99.75\% & 0.5 & 	99.81\% & 0.89 & 	19.14\% & \textbf{0.92}  \tabularnewline
\hline 
$semsim_{CF,ave}$   & 83.57\% & 0.66 & 89.22\% & 0.89 & 	88.35\% & 0.4 & 	97.79\% & 0.64 & 	95.96\% & 0.35 & 	99.55\% & 0.48 & 	99.67\% & 0.93 & 	14.43\% & 0.91 \tabularnewline
\hline 
$semsim_{TD,ave}$   & 77.83\% & 0.62 &  73.09\% & 0.84 & 	 87.39\% & 0.17 & 	 90.77\% & 0.67 & 	 83.32\% & 0.45 & 	 98.69\% & 0.47 & 	 99.21\% & 0.94 & 	 12.34\% & 0.78 \tabularnewline
\hline 
$semsim_{IIC,ave}$   & 79.67\% & 0.65 & 78.17\% & 0.85 & 	87.53\% & 0.26 & 	92.48\% & 0.71 & 	87.95\% & 0.43 & 	98.98\% & 0.5 & 	99.34\% & \textbf{0.95} & 	13.21\% & 0.82 \tabularnewline
\hline 
\hline 

$semsim_{AF,gav}$  & 85.67\% & \textbf{0.69} & 94.07\% & 0.94 & 	\textbf{90.21\%} & 0.52 & 	99.11\% & 0.69 & 	98.79\% & 0.45 & 	99.69\% & \textbf{0.52} & 	\textbf{99.84\%} & 0.83 & 	17.96\% & \textbf{0.92}  \tabularnewline
\hline 
$semsim_{CF,gav}$  & 83.43\% & \textbf{0.69} & 88.63\% & 0.93 & 	88.58\% & 0.41 & 	97.46\% & 0.72 & 	97.09\% & 0.5 & 	99.43\% & 0.5 & 	99.71\% & 0.88 & 	13.14\% & 0.91  \tabularnewline
\hline 
$semsim_{TD,gav}$  & 77.73\% & 0.65 &  71.36\% & 0.89 & 	 87.57\% & 0.18 & 	 89.32\% & 0.74 & 	 87.27\% & 0.58 & 	 98.3\% & 0.48 & 	 99.29\% & 0.92 & 	 11.0\% & 0.78  \tabularnewline
\hline 
$semsim_{IIC,gav}$  & 79.52\% & 0.68 & 76.75\% & 0.9 & 	87.7\% & 0.28 & 	91.28\% & \textbf{0.78} & 	91.01\% & 0.57 & 	98.69\% & \textbf{0.52} & 	99.41\% & 0.92 & 	11.83\% & 0.82  \tabularnewline
\hline 
\end{tabular} 

\vspace*{3cm} 

\caption{Degree of confidence \textbf{I1} and Pearson correlation \textbf{I2} values for the best performing \textit{SemSim$^p$} configuration, and the other selected methods. The table also includes the average values for both the indicators in the ``Average'' column.} 
\label{tab:compared_results_Spiss_ALL} 
\vspace*{0.1in} 
\tiny
\begin{tabular}{| c  | c | c || c | c | c | c | c | c | c | c | c | c | c | c | c | c | c | c | c | c | c | c | c | c | c | c | c | c | c | c | c | c |} 
\hline 
 &  \multicolumn{2}{c|}{Average} &  \multicolumn{2}{c|}{Sp.Iss.1} & \multicolumn{2}{c|}{Sp.Iss.2} & \multicolumn{2}{c|}{Sp.Iss.3} & \multicolumn{2}{c|}{Sp.Iss.4} & \multicolumn{2}{c|}{Sp.Iss.5} & \multicolumn{2}{c|}{Sp.Iss.6} & \multicolumn{2}{c|}{Sp.Iss.7}  \tabularnewline
\hline 
\textbf{Similarity method} &  {\textbf{I1}} & {\textbf{I2}} & 
  {\textbf{I1}} & {\textbf{I2}} &
   {\textbf{I1}} & {\textbf{I2}} &
   {\textbf{I1}} & {\textbf{I2}} &
   {\textbf{I1}} & {\textbf{I2}} &
   {\textbf{I1}} & {\textbf{I2}} &
   {\textbf{I1}} & {\textbf{I2}} &
   {\textbf{I1}} & {\textbf{I2}} 
 \tabularnewline 
\hline 
\hline 

$semsim_{AF,gav}$  & \textbf{85.67\%} & \textbf{0.69} & \textbf{94.07\%} & \textbf{0.94} & 	\textbf{90.21\%} & 0.52 & 	\textbf{99.11\%} & 0.69 & 	98.79\% & 0.45 & 	99.69\% & 0.52 & 	\textbf{99.84\%} & 0.83 & 	17.96\% & \textbf{0.92}  \tabularnewline
\hline 
\hline 
Dice & 84.76\% & 0.64 &  90.13\% & 0.69 & 86.53\% & 0.69 & 98.37\% & 0.72 & 96.92\% & 0.52 & 99.87\% & 0.73 & 99.51\% & 0.46 & 21.97\% & - \tabularnewline
\hline 
Jaccard & 82.37\% & 0.65 & 83.31\% & 0.72 & 78.23\% & \textbf{0.71} & 95.69\% & 0.73 & 95.89\% &0.45 & 99.77\% & \textbf{0.74} & 99.67\% & 0.58 & 	 24.05\% & -  \tabularnewline
\hline 
Sigmoid & 80.88\% & 0.63 &  80.84\% & 0.72  & 74.16\% & 0.67 & 92.02\% & 0.68 & 94.66\% & 0.45 & 99.4\% & 0.71 & 99.49\% & 0.58  & \textbf{25.56}\% & -   \tabularnewline
\hline
WNSim\textsubscript{sym} & 84.33\% & 0.61 &  92.0\% & 0.83  & 78.87\% & 0.69  & 98.79\% & 0.72  & 	 \textbf{98.92\%} & \textbf{0.59}  & 	\textbf{99.88\%} & 0.66 & 99.72\% & 0.35 & 	22.13\% & 0.46   \tabularnewline
\hline
Rezaei \& Fr{\"a}nti & 74.73\% & 0.46 &  60.09\% & 0.81  & 84.75\% & -0.06 & 79.81\% & 0.51 & 79.83\% & -0.08 & 94.91\% & 0.59 & 98.90\% & \textbf{0.84}  & 24.80\% & 0.61   \tabularnewline
\hline
Haase et al.\textsubscript{sym}  & 77.89\% & \textbf{0.69} & 68.94\% &  0.93 & 88.64\% & 0.28 & 81.91\% & \textbf{0.77} & 91.65\% & 0.7 & 96.92\% & 0.67 & 99.5\% &  0.82 & 17.65\% & 0.68   \tabularnewline  

\hline

\hline 
\end{tabular} 

\end{sidewaystable}

\begin{figure}[h]
	\includegraphics[width=\linewidth]{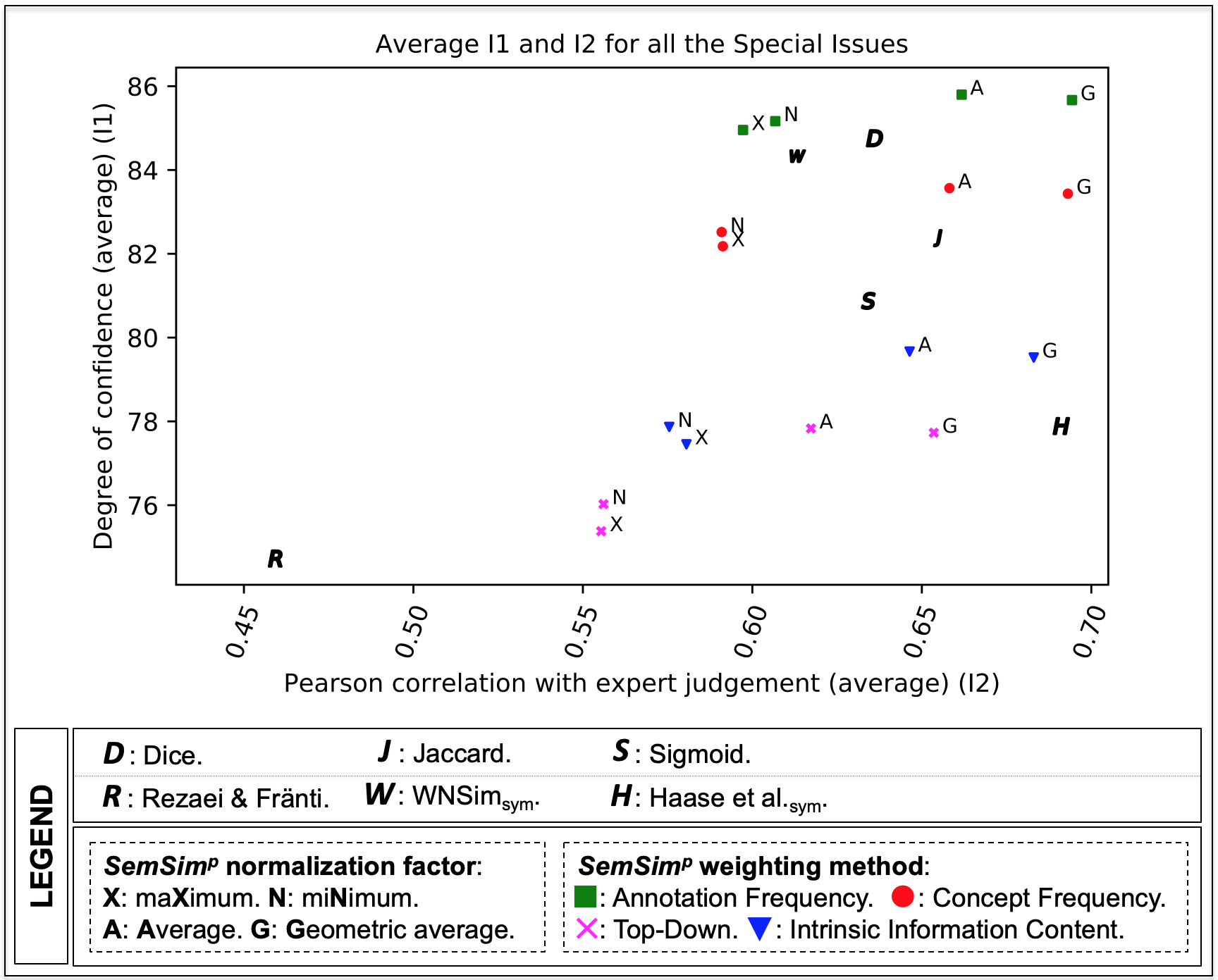}
	\caption{Scatter plot showing the average degree of confidence and the average Pearson correlation values computed for all the special issues and
	all the addressed methods.}
	\label{fig:assessment}
\end{figure}

\subsection{Threats to validity} \label{sect:threats}

This Section reports the approach that was adopted to cope with the threats to validity of the experiment in order to evaluate  the performance of an ontology-based similarity method. It relies on the notion of quality of a software engineering experiment that, according to Kaner \& Bond \cite{Kaner04softwareengineering} and Sj{\o}eberg et al. \cite{1514443}, depends on its \textbf{construct}, \textbf{internal} and \textbf{external} validities.

Threats to \textbf{construct validity} deal with limitations of the experimental set up. In particular, the experiment
could be limited by the general problem of \textit{lack of ontological quality} of the considered taxonomy, by the \textit{dataset size}, by the \textit{annotation quality}, and by the \textit{lack of availability of subject matter experts} for validation tasks. 

The lack of ontological quality concerns the syntactic, semantic, and social dimensions \citep{BURTONJONES200584}. 
In order to deal with it, we used the ontology derived from the ACM-CCS 2012. 
The syntactic quality is guaranteed by the {P}rot\'eg\'e ontology management system\footnote{See \url{https://protege.stanford.edu} \citep{musen2015the}.} 
that allowed us to validate the OWL code automatically generated from the ACM-CCS keywords.
Then the social and semantic dimensions of ontological quality are both guaranteed by the fact that the ontology was built on the basis of the ACM-CCS 2012, which is widely accepted and used in the computer science community.

Threats to dataset size deal with the choice of the appropriate size of the dataset. From one side, the risk is to choose a ``toy case study'' that does not reflect the complexity of the reality. From the other side, the risk is to have a too large dataset for which it is not feasible neither to ask experts to analyze it, nor to identify some criteria in order to select representative samples of data to be verified by expert judgement.
To reduce these risks, we used a large dataset including an ontology of 2,114 concepts and 1,103 annotated papers, and we selected the special issues from TOIS as representative samples to be evaluated by human experts.  

With regard to the annotation quality, as mentioned, in order to guarantee a good level of annotation, we selected two top quality journals, where most of the papers have been manually annotated by authors with keywords. Such annotations have also been checked by the reviewers of the papers and the editors of the journals and revised by the ACM in order to be compliant with the last release of the ACM-CCS (see Section \ref{sec:Studysettings}). 
Then we performed two expert judgement evaluations (i.e., one focusing on  keywords and the other also  on  abstracts) that allowed us to further discuss this issue in Section \ref{sect:discussion}.


The lack of availability of subject matter experts was addressed by considering a special issue as a ``cluster'' of mutually similar papers related to a specific topic (i.e., the special issue research theme). This assumption allowed us to perform the first experiment without involving human experts in evaluating the semantic similarity scores of a high number of pairs of papers. This is not the case of the second experiment, where the number of papers in each special issue was limited (i.e., 4 or 5) and it was quite easy to find experts available for the similarity evaluation as they were not overwhelmed in performing such a task.


\textbf{Internal validity} threats, instead, deal with the randomness of the results due to, for instance, the \textit{evaluator subjectiveness} and the \textit{explosion of the number of similarity values} to be considered that could lead to a burden of work affecting the quality of the evaluation. The first risk was reduced by designing the semantic cohesion experiment, centered on a statistical approach, where the human intervention was limited to the selection of papers done by the special issue editors.
However, since the editors may have not grouped the papers by similarity but, for instance, by discipline coverage, we did a further verification with the expert judgement experiment.
The second risk was reduced by asking experts to evaluate only the papers collected in the special issues. Indeed, they represent a limited but significant amount of samples of papers, considered similar by the authoritative opinion of the editors.

Finally, \textbf{external validity} threats concern the generalizability of the results. 
In this work, such a risk was reduced by addressing different topics in the computer science domain by selecting seven special issues.
In addition, a further experiment in the physical domain was performed, as illustrated in the next subsection.

\subsubsection{APS Experiment for External Validity} \label{sect:APSExperiment}


The domain of this experimentation concerns the papers published by the American Physical Society (APS) in the Physical Review Letters, a journal covering the full arc of fundamental and interdisciplinary physics research. In the experiment, we considered 52,762  articles published in this journal from 1980 to 2013, and annotated according to the Physics and Astronomy Classification Scheme (PACS), a classification system of the APS, organized as a taxonomy. There are several updated editions of the PACS until 2010. Since 2016, the PACS has been replaced by the PhySH (Physics Subject Headings). In this paper, we consider the PACS 2010 regular edition,  consisting of 4,574 subjects covering all the fields of physics. 


Analogously to the statistical part of the ACM experiment, we evaluated the performance of the 16 different configurations of \textit{SemSim$^p$}, Dice, Jaccard, Sigmoid, WNSim similarity methods and those proposed, respectively, by Haase et al., and Rezaei \& Fr{\"a}nti. In particular, we computed the degree of confidence \textbf{I1} in order to answer to \textbf{RQ1}.


Concerning the selection of special issues, similar to the ACM experiment,
we considered a collection of 21 papers regarding high-temperature superconductors, which was assembled by the APS editorial office\footnote{See \url{https://journals.aps.org/prl/heating-up-of-superconductors}. Last access on 16\textsuperscript{th} April, 2022.}. 
However, randomly sampling 21 mutually similar papers out of 52,762 multidisciplinary ones is an extremely rare event.
Hence, the likelihood that the semantic cohesion of the 21 mentioned papers is higher than the one of 100,000 sets of randomly selected papers with the same cardinality is close to 100\%  for every method. 
For this reason, in line with the statistical experiment performed on the ACM dataset, we selected 4 papers from the mentioned collection.
 For the sake of clarity, in the rest of this paper, we refer to this collection of 4 papers as the \textit{APS virtual special issue}.


We built the reference ontology in accordance with the structure and format of the PACS codes\footnote{See \url{https://web.archive.org/web/20131122200802/http://www.aip.org/pacs/pacs2010/about.html}}. It gathers 4,575 concepts and 4,574 ISA relationships,  and it has a tree-shaped structure with 6 levels. 
Analogously to the ACM dataset, we evaluated the weights  of the APS concepts according to the CF, AF, TD, and IIC weighting methods. The dataset used to compute the weights for the extensional methods (i.e., CF and AF) consists of all the 52,762 papers. 
Conversely, we restricted the statistical experiment to 2,000 randomly selected papers from this dataset, which include the 4 papers of the APS virtual special issue.   


Finally, we computed the \textit{t-value} and the corresponding degree of confidence (\textbf{I1}) of the asserted hypothesis, i.e., the papers of the APS virtual special issue are similar.
Table \ref{tab:APS_results_ALL} shows the results of this experiment for all the methods ordered from the highest to the lowest values. 

\begin{table*}[h] 
\centering 
\caption{Degree of confidence (\textbf{I1}) for the APS virtual special issue obtained by using all the methods.}
\label{tab:APS_results_ALL} 
\vspace*{0.1in} 
\tiny
\begin{tabular}{| c  | c |} 
\hline 
\textbf{Similarity method} &  \textbf{I1}  \tabularnewline 
\hline 
\hline 

$semsim_{AF,max}$  & \textbf{99.86\%}  \tabularnewline 
\hline 
$semsim_{TD,max}$  & \textbf{99.86\%}   \tabularnewline
\hline 
$semsim_{CF,max}$  & 99.84\%   \tabularnewline 
\hline 
$semsim_{IIC,max}$  & 99.84\%    \tabularnewline
\hline 
$semsim_{TD,ave}$   & 99.84\%  \tabularnewline
\hline 
$semsim_{AF,ave}$   & 99.83\%  \tabularnewline
\hline 
Haase et al.\textsubscript{sym}  & 99.83\%  \tabularnewline  
\hline 
$semsim_{CF,ave}$   & 99.82\%  \tabularnewline
\hline 
$semsim_{TD,gav}$  & 99.82\%   \tabularnewline
\hline 
$semsim_{AF,gav}$  & 99.82\%  \tabularnewline
\hline 
$semsim_{IIC,ave}$   & 99.81\% \tabularnewline
\hline 
$semsim_{CF,gav}$  & 99.80\%   \tabularnewline
\hline 
$semsim_{IIC,gav}$  & 99.79\%  \tabularnewline
\hline
Rezaei \& Fr{\"a}nti & 99.79\%  \tabularnewline
\hline 
$semsim_{TD,min}$  & 99.70\%  \tabularnewline
\hline 
$semsim_{AF,min}$   & 99.69\%   \tabularnewline
\hline 
$semsim_{CF,min}$  & 99.66\%   \tabularnewline
\hline 
$semsim_{IIC,min}$   & 99.65\%  \tabularnewline
\hline 
WNSim\textsubscript{sym} & 97.91\%  \tabularnewline
\hline
Dice & 97.39\%   \tabularnewline
\hline 
Jaccard & 96.24\%    \tabularnewline
\hline 
Sigmoid & 95.30\%  \tabularnewline
\hline
\end{tabular} 
\centering 
\end{table*}

\section{Discussion} \label{sect:discussion}
\subsection{ACM experiments}\label{sec:ACMexpDiscussion}

In Figure \ref{fig:assessment}, as mentioned in Section \ref{Sect:studyresults}, 
the scatter plot shows the average degree of confidence and the average Pearson correlation values computed for all the seven special issues, and for all the addressed methods.

Note that, concerning {\bf I1}, \textit{SemSim$^p$} configurations with the extensional weighting methods ($AF$ and $CF$) provide better results with respect to those with the intensional ones ($TD$ and $IIC$) and, in particular,  $AF$  shows the highest degree of confidence. Furthermore, independently of the weighting methods, the configurations with $A$ (arithmetic average), as similarity normalization factor, provide the best degree of confidence values, although they are very close to those with $G$ (geometric average). We also observe that all the \textit{SemSim$^p$} configurations with $AF$ outperform the other selected methods.

With regard to  {\bf I2}, the \textit{SemSim$^p$} configuration with $AF$, as weighting method, and $G$, as similarity normalization factor, shows better results with respect to the other configurations and all the selected methods except for Haase et al.\textsubscript{sym}, which has a comparable {\bf I2} value.
 Furthermore, given a \textit{SemSim$^p$} weighting method, the configuration with $G$  provides the best average Pearson correlation values, and significantly outperforms that with $A$.

The study also reveals that, by focusing on  the degrees of confidence \textbf{I1} of \textit{SemSim$^p$} with the G and A similarity normalization factors, the results according to  the AF weighting method are very close (0.13$\%$) whereas, according to the indicator \textbf{I2}, they differ significantly (3$\%$). 

Overall, given the above mentioned results, in our opinion the \textit{SemSim$^p$} configuration with the $AF$ weighting method and the $G$ similarity normalization factor is recommended. In fact, the configuration with $A$ slightly outperforms that with $G$ with respect to  {\bf I1} (but such a difference  can be neglected), whereas  the difference between $G$ and $A$ configurations with respect to {\bf I2} is relevant and cannot be ignored.

\begin{figure}[h]
	\centering
	\includegraphics[width=\linewidth]{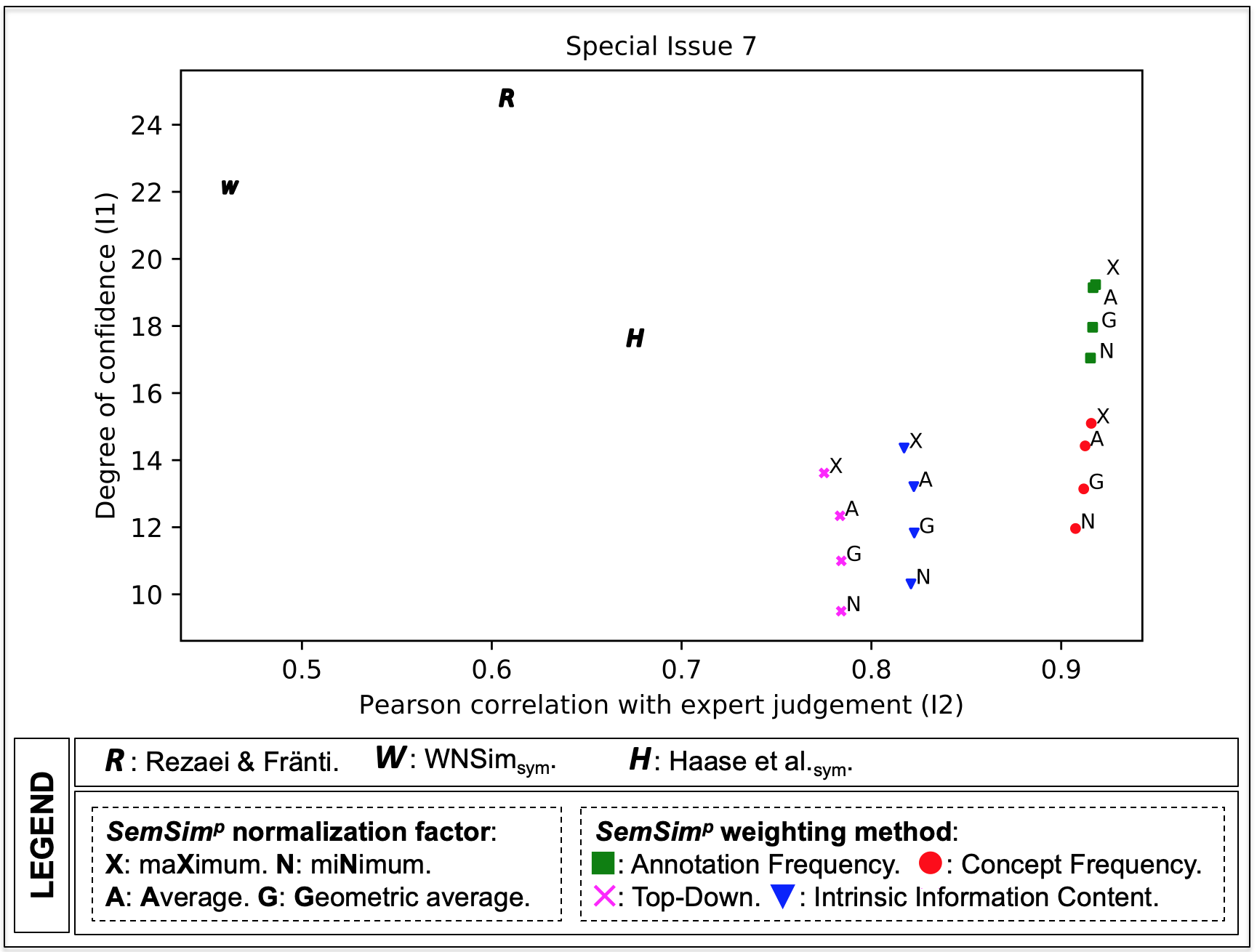}
	\caption{Scatter plot concerning the average degree of confidence and the average Pearson correlation values for all the \textit{SemSim$^p$} configurations, WNSim\textsubscript{sym}, Haase et al.\textsubscript{sym}, and the method of Rezaei \& Fr{\"a}nti concerning the Special Issue 7.}
	\label{fig:SpIss_7DJ}
\end{figure}

Finally, Figure \ref{fig:SpIss_7DJ} shows the average degrees of confidence and the average Pearson correlation values in a scatter plot related to all the \textit{SemSim$^p$} configurations,  WNSim\textsubscript{sym}, Haase et al.\textsubscript{sym}, and the method of Rezaei \& Fr{\"a}nti  for the Special Issue 7. 
In this case, the degrees of confidence of \textit{SemSim$^p$} configurations are low and, hence, the {\bf RQ1} assumption (see Section \ref{sec:Research_Questions}), related to the assembly of the papers by following some similarity criteria, seems not to be satisfied.
This is also confirmed by the semantic cohesion related to keywords ({\it EJ\textsubscript{key}} in Table 
\ref{tab:ave_sim_values}), 
which is the lowest value (0.18) among the others.
We believe that this is due to the low quality of the annotation for this specific special issue (see threats to construct validity in Section \ref{sect:threats}). 
In fact, the semantic cohesion value increases (0.39) in the case it is estimated by also considering the abstracts ({\it EJ\textsubscript{abs}}), because the evaluation has been performed according to a richer amount of information. 
Indeed, as shown in Table 
\ref{tab:ave_sim_values},
 this is evident by {looking at the difference  (0.21) between the semantic cohesion  related to {\it EJ\textsubscript{key}} and the one related to the {\it EJ\textsubscript{abs}}, which is greater with respect to the other cases.
The reason of the low quality of the annotation is that in 2 out of 5  papers of the Special Issue 7, the selected keywords are too generic (e.g., Information Retrieval, see Figure \ref{fig:acm_excerpt}), therefore  they do not  adequately describe the contents of the papers. Furthermore, as shown in Table \ref{tab:special_issue_data_II}, in general the papers of the Special Issue 7 have been annotated with fewer keywords. This also applies to the Special Issue 6, but in this case the selected keywords are hierarchically related (they belong to the same paths of the ISA hierarchy). Therefore the {\bf RQ1} assumption holds for the Special Issue 6, as well as the special issues 1-5.


\subsection{The APS experiment}\label{sec:APSexpDiscussion}

The experiment with APS is different with respect to the ACM ones. Firstly, it concerns a different application domain, i.e., physics. Then, within this field, the considered papers cover the whole range of physics and are therefore less focused than the topics addressed by the ACM TODS and TOIS journals, which mainly concern information systems. Accordingly, as shown in Table \ref{tab:APS_results_ALL}, the degree of confidence of the asserted hypothesis (\textbf{I1}) is very high for all the assessed methods, since it is quite easy to identify the virtual special issue among the sets of randomly sampled papers.
Our study reveals that all \textit{SemSim$^p$} configurations significantly outperform the WNSim\textsubscript{sym}, Dice, Jaccard, and Sigmoid similarity methods.
The \textit{SemSim$^p$} configurations with M (maximum), A, and G, together with the Haase et al.\textsubscript{sym} and Rezaei \& Fr{\"a}nti methods are the best performing ones, and almost equivalent because their  difference is less than 0.07\%.

In general, the APS experiment confirms that \textit{SemSim$^p$} provides the best results. The main difference with respect to the ACM experimentation is that the choice of the
weighting method plays a less relevant role in determining the best method. This is probably due to the above-mentioned inherent characteristics of the APS articles, which are less focused than the ACM ones. 
This is because the 52,762 papers cover all the fields of the physics and, then, when weights are computed with the extensional methods, i.e, AF and CF, they are comparable to those obtained according to the intensional ones, i.e, TD and IIC.

\section{Conclusion} \label{sect:conclusion}

In this work, the parametric semantic similarity method \textit{SemSim$^p$} has been introduced. It is based  on a reference ontology, which is used to semantically annotate a set of digital resources. 
This method was successfully validated
by means of a comparative assessment, which was performed through an experimentation that includes both a statistical analysis and an expert judgement evaluation. To this end, we adopted a large dataset of 1,103 papers collected from the Digital Library of the ACM and an ontology derived from the ACM Computing Classification System, which is one of the standard classification systems in the computing field. 
Then, we adopted the APS dataset in the domain of physics to assess the generalizability of the results.

For the ACM dataset, where the annotated digital resources mainly focus on one subject, i.e., information systems (see Section \ref{sec:APSexpDiscussion}), 
the comparative assessment
 allowed us to elaborate the following considerations about the \textit{SemSim$^p$} configurations. 
In particular, the one with the annotation frequency weighting method (AF) and the geometric average similarity normalization factor (\textit{gav}) is recommended, since it exhibits the highest correlation with expert judgement and a high degree of confidence. 
However, for the datasets for which a significant amount of annotated resources is not available, \textit{SemSim$^p$} allows the use of the intensional methods.
In particular, when it is configured with IIC and the geometric average similarity normalization factor (i.e., \textit{gav}), it still provides results comparable to the best performing methods (i.e., Dice, Jaccard, WNSim\textsubscript{sym}, Sigmoid, and Haase et al.\textsubscript{sym}). In fact, Dice, WNSim\textsubscript{sym}, Jaccard, and Sigmoid outperform this \textit{SemSim$^p$}  configuration with regard to \textbf{I1}, but it is not the case with regard to \textbf{I2}. Conversely, this \textit{SemSim$^p$} configuration outperforms Haase et al.\textsubscript{sym} with regard to \textbf{I1}, but it is slightly outperformed with regard to \textbf{I2}.

We also assessed the adoption of different normalization factors with the aim of improving \textit{SemSim} \citep{FMPT13}. Indeed, the presented analysis reveals that the best performance of \textit{SemSim$^p$} is achieved with the \textit{ave} and the \textit{gav} similarity normalization factors.

Note that, in this work, we addressed the problem of expert judgement related to  the difficulty of asking a significant number of experts to pairwise check thousands of resources, producing millions of similarity scores. In particular, we adopted a new solution to assess similarity methods, which relies on a large dataset consisting of 1,103 annotated papers and seven ACM special issues, and the assumption that the papers published in a special issue are gathered by the editors according to a given research topic indicated in the call for paper. 

The APS experimentation, where the annotated digital resources are less focused, reveals that, in general, \textit{SemSim$^p$} provides better results than other similarity methods 
but the choice of the weighting method is less relevant.

As future work, we plan to experiment \textit{SemSim$^p$} in the health sector in order to further investigate its applicability to different contexts, and define the guidelines to identify the best \textit{SemSim$^p$} configuration depending on the specificities of the collected dataset of a given domain.




\section*{Acknowledgement}
We wish to acknowledge the contribution that our colleague Elaheh Pourabbas provided in an important part of the paper. Unfortunately, a premature passing away prevented her to enjoy the final achievement of the work with the publication of the paper on this journal. We want to remember our departed colleague and preserve her memory among the dearest ones.

We thank the American Physical Society for providing us with the dataset. Furthermore, we acknowledge the Association for Computing Machinery, whose Computing Classification System enabled us to carry out the experiments presented in this article.



\bibliographystyle{plainnat}  
\bibliography{elsarticle-template}

\begin{thebibliography}{67}
\providecommand{\natexlab}[1]{#1}
\providecommand{\url}[1]{\texttt{#1}}
\expandafter\ifx\csname urlstyle\endcsname\relax
  \providecommand{\doi}[1]{doi: #1}\else
  \providecommand{\doi}{doi: \begingroup \urlstyle{rm}\Url}\fi

\bibitem[Abdelaziz et~al.(2017)Abdelaziz, Fokoue, Hassanzadeh, Zhang, and
  Sadoghi]{Abdelaziz2017}
Ibrahim Abdelaziz, Achille Fokoue, Oktie Hassanzadeh, Ping Zhang, and Mohammad
  Sadoghi.
\newblock Large-scale structural and textual similarity-based mining of
  knowledge graph to predict drug-drug interactions.
\newblock \emph{Journal of Web Semantics}, 44:\penalty0 104--117, 2017.
\newblock \doi{10.1016/j.websem.2017.06.002}.
\newblock URL \url{https://doi.org/10.1016/j.websem.2017.06.002}.

\bibitem[Abioui et~al.(2018)Abioui, Idarrou, Bouzit, and Mammass]{Abioui2018}
Hasna Abioui, Ali Idarrou, Ali Bouzit, and Driss Mammass.
\newblock Towards a novel and generic approach for {OWL} ontology weighting.
\newblock \emph{Procedia Computer Science}, 127:\penalty0 426 -- 435, 2018.
\newblock ISSN 1877-0509.
\newblock \doi{10.1016/j.procs.2018.01.140}.

\bibitem[Bastian et~al.(2009)Bastian, Heymann, and Jacomy]{ICWSM09154}
Mathieu Bastian, Sebastien Heymann, and Mathieu Jacomy.
\newblock Gephi: An open source software for exploring and manipulating
  networks.
\newblock In \emph{Proceedings of the International AAAI Conference on Weblogs
  and Social Media}, 2009.
\newblock URL \url{http://www.aaai.org/ocs/index.php/ICWSM/09/paper/view/154}.

\bibitem[Bloehdorn and Moschitti(2007)]{Bloehdorn2007}
Stephan Bloehdorn and Alessandro Moschitti.
\newblock Combined syntactic and semantic kernels for text classification.
\newblock In Giambattista Amati, Claudio Carpineto, and Giovanni Romano,
  editors, \emph{Advances in Information Retrieval, 29th European Conference on
  {IR} Research, {ECIR} 2007, Rome, Italy, April 2-5, 2007, Proceedings},
  volume 4425 of \emph{Lecture Notes in Computer Science}, pages 307--318.
  Springer, 2007.
\newblock \doi{10.1007/978-3-540-71496-5\_29}.
\newblock URL \url{https://doi.org/10.1007/978-3-540-71496-5\_29}.

\bibitem[Bluman(2009)]{bluman2009elementary}
Allan~G. Bluman.
\newblock \emph{Elementary statistics: A step by step approach}.
\newblock McGraw-Hill Higher Education New York, 2009.

\bibitem[Bogdanovic et~al.(2021)Bogdanovic, Veljkovic, Gligorijevic, Puflovic,
  and Stoimenov]{Bogdanovic2021}
Milos Bogdanovic, Natasa Veljkovic, Milena~Frtunic Gligorijevic, Darko
  Puflovic, and Leonid Stoimenov.
\newblock On revealing shared conceptualization among open datasets.
\newblock \emph{Journal of Web Semantics}, 66:\penalty0 100624, 2021.
\newblock \doi{10.1016/j.websem.2020.100624}.
\newblock URL \url{https://doi.org/10.1016/j.websem.2020.100624}.

\bibitem[Burton-Jones et~al.(2005)Burton-Jones, Storey, Sugumaran, and
  Ahluwalia]{BURTONJONES200584}
Andrew Burton-Jones, Veda~C. Storey, Vijayan Sugumaran, and Punit Ahluwalia.
\newblock A semiotic metrics suite for assessing the quality of ontologies.
\newblock \emph{Data \& Knowledge Engineering}, 55\penalty0 (1):\penalty0 84 --
  102, 2005.
\newblock ISSN 0169-023X.
\newblock \doi{10.1016/j.datak.2004.11.010}.

\bibitem[Chandrasekaran and Mago(2021)]{Chandra2021}
Dhivya Chandrasekaran and Vijay Mago.
\newblock {Evolution of Semantic Similarity - A Survey}.
\newblock \emph{ACM Computing Survey}, 54\penalty0 (2), February 2021.
\newblock ISSN 0360-0300.

\bibitem[Chen et~al.(2017)Chen, Lu, Wu, and Li]{Chen201719}
F.~Chen, C.~Lu, H.~Wu, and M.~Li.
\newblock A semantic similarity measure integrating multiple conceptual
  relationships for web service discovery.
\newblock \emph{Expert Systems with Applications}, 67:\penalty0 19--31, 2017.
\newblock ISSN 09574174.
\newblock \doi{10.1016/j.eswa.2016.09.028}.

\bibitem[Chen et~al.(2020)Chen, Wang, Zhao, Yin, Markov, and
  Rijke]{10.1145/3372154}
Yifan Chen, Yang Wang, Xiang Zhao, Hongzhi Yin, Ilya Markov, and Maarten~De
  Rijke.
\newblock Local variational feature-based similarity models for recommending
  top-n new items.
\newblock \emph{ACM Transactions on Information Systems}, 38\penalty0 (2),
  2020.
\newblock ISSN 1046-8188.
\newblock \doi{10.1145/3372154}.
\newblock URL \url{https://doi.org/10.1145/3372154}.

\bibitem[Cord{\`{\i}} et~al.(2005)Cord{\`{\i}}, Lombardi, Martelli, and
  Mascardi]{CordiLMM05}
Valentina Cord{\`{\i}}, Paolo Lombardi, Maurizio Martelli, and Viviana
  Mascardi.
\newblock An ontology-based similarity between sets of concepts.
\newblock In Flavio Corradini, Flavio~De Paoli, Emanuela Merelli, and Andrea
  Omicini, editors, \emph{{WOA} 2005: Dagli Oggetti agli Agenti. 6th
  AI*IA/TABOO Joint Workshop ``From Objects to Agents'': Simulation and Formal
  Analysis of Complex Systems, 14-16 November 2005, Camerino, MC, Italy}, pages
  16--21. Pitagora Editrice Bologna, 2005.
\newblock URL \url{http://lia.deis.unibo.it/books/woa2005/papers/3.pdf}.

\bibitem[d'Amato(2007)]{DAmato2007}
Claudia d'Amato.
\newblock \emph{Similarity-based Learning Methods for the Semantic Web}.
\newblock PhD thesis, Universit{{\'a}}  degli Studi di Bari, Facolt{{\'a}} di
  Scienze, Dipartimento di Informatica, 2007.

\bibitem[De~Nicola and D'Agostino(2021)]{DeNicola2020}
Antonio De~Nicola and Gregorio D'Agostino.
\newblock Assessment of gender divide in scientific communities.
\newblock \emph{Scientometrics}, May 2021:\penalty0 3807 -- 3840, 2021.
\newblock \doi{10.1007/s11192-021-03885-3}.

\bibitem[De~Nicola and Missikoff(2016)]{denicola2016}
Antonio De~Nicola and Michele Missikoff.
\newblock A lightweight methodology for rapid ontology engineering.
\newblock \emph{Communications of the ACM}, 59\penalty0 (3):\penalty0 79--86,
  2016.
\newblock \doi{10.1145/2818359}.

\bibitem[{De Nicola} and Taglino(2023)]{https://doi.org/10.17632/pv3hvmbppk.1}
Antonio {De Nicola} and Francesco Taglino.
\newblock {SemSimp} similarity experiments - software and data, 2023.
\newblock URL \url{https://data.mendeley.com/datasets/pv3hvmbppk/1}.

\bibitem[De~Nicola et~al.(2009)De~Nicola, Missikoff, and Navigli]{denicola2009}
Antonio De~Nicola, Michele Missikoff, and Roberto Navigli.
\newblock A software engineering approach to ontology building.
\newblock \emph{Information systems}, 34\penalty0 (2):\penalty0 258--275, 2009.
\newblock \doi{10.1016/j.is.2008.07.002}.

\bibitem[{De Nicola} et~al.(2019){De Nicola}, Formica, Missikoff, Pourabbas,
  and Taglino]{DFMPT19}
Antonio {De Nicola}, Anna Formica, Michele Missikoff, Elaheh Pourabbas, and
  Francesco Taglino.
\newblock A comparative assessment of ontology weighting methods in semantic
  similarity search.
\newblock In Ana~Paula Rocha, Luc Steels, and H.~Jaap van~den Herik, editors,
  \emph{Proceedings of the 11th International Conference on Agents and
  Artificial Intelligence, {ICAART} 2019, Volume 2, Prague, Czech Republic,
  February 19-21, 2019}, pages 506--513. SciTePress, 2019.
\newblock \doi{10.5220/0007342805060513}.

\bibitem[Dice(1945)]{Dice1945}
Lee~R. Dice.
\newblock Measures of the amount of ecologic association between species.
\newblock \emph{Ecology}, 26\penalty0 (3):\penalty0 297--302, 1945.
\newblock \doi{10.2307/1932409}.

\bibitem[Dulmage and Mendelsohn(1958)]{Dulmage}
A.~L. Dulmage and Nathan~Saul Mendelsohn.
\newblock Coverings of bipartite graphs.
\newblock \emph{Canadian Journal of Mathematics}, 10:\penalty0 517--534, 1958.
\newblock \doi{10.4153/CJM-1958-052-0}.

\bibitem[Espinoza-Arias et~al.(2021)Espinoza-Arias, Garijo, and
  Corcho]{ESPINOZAARIAS2021100655}
Paola Espinoza-Arias, Daniel Garijo, and Oscar Corcho.
\newblock Crossing the chasm between ontology engineering and application
  development: A survey.
\newblock \emph{Journal of Web Semantics}, 70:\penalty0 100655, 2021.
\newblock ISSN 1570-8268.
\newblock \doi{10.1016/j.websem.2021.100655}.
\newblock URL
  \url{https://www.sciencedirect.com/science/article/pii/S1570826821000305}.

\bibitem[Formica and Taglino(2021)]{Formica2021}
Anna Formica and Francesco Taglino.
\newblock An enriched information-theoretic definition of semantic similarity
  in a taxonomy.
\newblock \emph{{IEEE} Access}, 9:\penalty0 100583--100593, 2021.
\newblock \doi{10.1109/ACCESS.2021.3096598}.
\newblock URL \url{https://doi.org/10.1109/ACCESS.2021.3096598}.

\bibitem[Formica et~al.(2008)Formica, Missikoff, Pourabbas, and
  Taglino]{FMPT08}
Anna Formica, Michele Missikoff, Elaheh Pourabbas, and Francesco Taglino.
\newblock Weighted ontology for semantic search.
\newblock In \emph{Proc. of the OTM 2008 Confederated International
  Conferences, CoopIS, DOA, GADA, IS, and ODBASE 2008. Part II on On the Move
  to Meaningful Internet Systems}, OTM '08, pages 1289--1303, Berlin,
  Heidelberg, 2008. Springer-Verlag.
\newblock ISBN 978-3-540-88872-7.
\newblock \doi{10.1007/978-3-540-88873-4_26}.

\bibitem[Formica et~al.(2013)Formica, Missikoff, Pourabbas, and
  Taglino]{FMPT13}
Anna Formica, Michele Missikoff, Elaheh Pourabbas, and Francesco Taglino.
\newblock Semantic search for matching user requests with profiled enterprises.
\newblock \emph{Computers in Industry}, 64\penalty0 (3):\penalty0 191--202,
  April 2013.
\newblock ISSN 0166-3615.
\newblock \doi{10.1016/j.compind.2012.09.007}.

\bibitem[Formica et~al.(2016)Formica, Missikoff, Pourabbas, and
  Taglino]{FMPT16}
Anna Formica, Michele Missikoff, Elaheh Pourabbas, and Francesco Taglino.
\newblock A {B}ayesian approach for weighted ontologies and semantic search.
\newblock In \emph{Proc. of the 8th Int. Joint Conf. on Knowledge Discovery,
  Knowledge Engineering and Knowledge Management {(IC3K} 2016) - KEOD, Porto -
  Portugal, November 9 - 11, 2016.}, pages 171--178, 2016.
\newblock \doi{10.5220/0006073301710178}.

\bibitem[Formica et~al.(2020)Formica, Mazzei, Pourabbas, and
  Rafanelli]{9086014}
Anna Formica, Mauro Mazzei, Elaheh Pourabbas, and Maurizio Rafanelli.
\newblock Approximate query answering based on topological neighborhood and
  semantic similarity in openstreetmap.
\newblock \emph{IEEE Access}, 8:\penalty0 87011--87030, 2020.
\newblock \doi{10.1109/ACCESS.2020.2992202}.

\bibitem[Haase et~al.(2004)Haase, Siebes, and van Harmelen]{Hasse2004}
Peter Haase, Ronny Siebes, and Frank van Harmelen.
\newblock Peer selection in peer-to-peer networks with semantic topologies.
\newblock In Mokrane Bouzeghoub, Carole Goble, Vipul Kashyap, and Stefano
  Spaccapietra, editors, \emph{Semantics of a Networked World. Semantics for
  Grid Databases. ICSNW 2004. Lecture Notes in Computer Science, vol 3226},
  pages 108--125, Berlin, Heidelberg, 2004. Springer Berlin Heidelberg.
\newblock ISBN 978-3-540-30145-5.
\newblock \doi{10.1007/978-3-540-30145-5_7}.

\bibitem[Hassan et~al.(2019)Hassan, AbdelRahman, Bahgat, and Farag]{Hassan2019}
Basma Hassan, Samir~E. AbdelRahman, Reem Bahgat, and Ibrahim Farag.
\newblock {UESTS: An Unsupervised Ensemble Semantic Textual Similarity Method}.
\newblock \emph{{IEEE} Access}, 7:\penalty0 85462--85482, 2019.
\newblock \doi{10.1109/ACCESS.2019.2925006}.
\newblock URL \url{https://doi.org/10.1109/ACCESS.2019.2925006}.

\bibitem[Hassanpour et~al.(2014)Hassanpour, O'Connor, and Das]{Hassanpour2014}
Saeed Hassanpour, Martin~J. O'Connor, and Amar~K. Das.
\newblock Clustering rule bases using ontology-based similarity measures.
\newblock \emph{Journal of Web Semantics}, 25:\penalty0 1--8, 2014.
\newblock \doi{10.1016/j.websem.2014.03.001}.
\newblock URL \url{https://doi.org/10.1016/j.websem.2014.03.001}.

\bibitem[Hayuhardhika et~al.(2013)Hayuhardhika, Purta, Sugiyanto, Riyanarto,
  and Sidiq]{Hayuhardhika2013}
Widhy Hayuhardhika, Nugraha Purta, Sugiyanto, Sarno Riyanarto, and Mohamad
  Sidiq.
\newblock Weighted ontology and weighted tree similarity algorithm for
  diagnosing diabetes mellitus.
\newblock In \emph{2013 International Conference on Computer, Control,
  Informatics and Its Applications (IC3INA)}, pages 267--272, Nov 2013.
\newblock \doi{10.1109/IC3INA.2013.6819185}.

\bibitem[Jaccard(1912)]{Jaccard1912}
Paul Jaccard.
\newblock The distribution of the flora in the alpine zone.
\newblock \emph{New Phytologist}, 11:\penalty0 37--50, 1912.
\newblock \doi{10.1111/j.1469-8137.1912.tb05611.x}.

\bibitem[Jia et~al.(2019)Jia, Lu, Duan, and Li]{JiaLDL19}
Zheng Jia, Xudong Lu, Huilong Duan, and Haomin Li.
\newblock Using the distance between sets of hierarchical taxonomic clinical
  concepts to measure patient similarity.
\newblock \emph{{BMC} Medical Informatics and Decision Making}, 19\penalty0
  (1):\penalty0 91:1--91:11, 2019.
\newblock \doi{10.1186/s12911-019-0807-y}.

\bibitem[Jiang and Conrath(1997)]{Jiang1997}
Jay~J. Jiang and David~W. Conrath.
\newblock Semantic similarity based on corpus statistics and lexical taxonomy.
\newblock In \emph{Proceedings of the International Conference on Research in
  Computational Linguistics}, pages 19--33, 1997.
\newblock URL
  \url{http://www.cse.iitb.ac.in/~cs626-449/Papers/WordSimilarity/4.pdf}.

\bibitem[Kalos and Whitlock(2009)]{kalos2009monte}
Malvin~H. Kalos and Paula~A. Whitlock.
\newblock \emph{{Monte Carlo} methods}.
\newblock John Wiley \& Sons, 2009.

\bibitem[Kaner and Bond(2004)]{Kaner04softwareengineering}
Cem Kaner and Walter~P. Bond.
\newblock Software engineering metrics: What do they measure and how do we
  know?
\newblock In \emph{METRICS 2004. IEEE CS}. Press, 2004.

\bibitem[K{\"o}hler et~al.(2009)K{\"o}hler, Schulz, Krawitz, Bauer, D{\"o}lken,
  Ott, Mundlos, Horn, Mundlos, and Robinson]{kohler2009clinical}
Sebastian K{\"o}hler, Marcel~H Schulz, Peter Krawitz, Sebastian Bauer, Sandra
  D{\"o}lken, Claus~E Ott, Christine Mundlos, Denise Horn, Stefan Mundlos, and
  Peter~N Robinson.
\newblock Clinical diagnostics in human genetics with semantic similarity
  searches in ontologies.
\newblock \emph{The American Journal of Human Genetics}, 85\penalty0
  (4):\penalty0 457--464, 2009.

\bibitem[Leacock and Chodorow(1998)]{Leacock1998}
Claudia Leacock and Martin Chodorow.
\newblock Combining local context and wordnet similarity for word sense
  identification.
\newblock \emph{WordNet: An electronic lexical database, MIT Press},
  49:\penalty0 265--283, 1998.
\newblock ISSN 9780262272551.
\newblock \doi{10.7551/mitpress/7287.003.0018}.

\bibitem[Li et~al.(2006)Li, McLean, Bandar, O'Shea, and Crockett]{Li2006}
Yuhua Li, David McLean, Zuhair~A. Bandar, James~D. O'Shea, and Keeley Crockett.
\newblock Sentence similarity based on semantic nets and corpus statistics.
\newblock \emph{IEEE Transactions on Knowledge and Data Engineering},
  18\penalty0 (8):\penalty0 1138--1150, 2006.
\newblock \doi{10.1109/TKDE.2006.130}.

\bibitem[Likavec et~al.(2019)Likavec, Lombardi, and Cena]{LIKAVEC2019}
Silvia Likavec, Ilaria Lombardi, and Federica Cena.
\newblock Sigmoid similarity - a new feature-based similarity measure.
\newblock \emph{Information Sciences}, 481:\penalty0 203--218, 2019.
\newblock ISSN 0020-0255.
\newblock \doi{10.1016/j.ins.2018.12.018}.
\newblock URL
  \url{https://www.sciencedirect.com/science/article/pii/S0020025518309630}.

\bibitem[Lin(1998)]{Lin}
Dekang Lin.
\newblock An information-theoretic definition of similarity.
\newblock In \emph{Proceedings of the 15th International Conference on Machine
  Learning}, ICML '98, pages 296--304, San Francisco, CA, USA, 1998. Morgan
  Kaufmann Publishers Inc.
\newblock ISBN 1-55860-556-8.
\newblock URL \url{http://dl.acm.org/citation.cfm?id=645527.657297}.

\bibitem[Liu et~al.(2021)Liu, Wang, Li, Li, Deng, and Pan]{Liu2021}
Jinshuo Liu, Chenyang Wang, Chenxi Li, Ningxi Li, Juan Deng, and Jeff~Z. Pan.
\newblock {DTN: Deep triple network for topic specific fake news detection}.
\newblock \emph{Journal of Web Semantics}, 70:\penalty0 100646, 2021.
\newblock ISSN 1570-8268.
\newblock \doi{10.1016/j.websem.2021.100646}.
\newblock URL
  \url{https://www.sciencedirect.com/science/article/pii/S1570826821000214}.

\bibitem[Mandeep and Harries(2001)]{Mandeep2001}
K.~Dhami Mandeep and Clare Harries.
\newblock Fast and frugal versus regression models of human judgement.
\newblock \emph{Thinking \& Reasoning}, 7\penalty0 (1):\penalty0 5--27, 2001.
\newblock \doi{https://doi.org/10.1080/13546780042000019}.

\bibitem[Manning et~al.(2008)Manning, Raghavan, and Schutze]{Manning2008}
Christopher~D. Manning, Prabhakar Raghavan, and Hinrich Schutze.
\newblock \emph{Introduction to Information Retrieval}.
\newblock Cambridge University Press, New York, NY, USA, 2008.
\newblock ISBN 0521865719, 9780521865715.

\bibitem[Meng et~al.(2012)Meng, Gu, and Zhou]{Zhou2012}
Lingling Meng, Junzhong Gu, and Zili Zhou.
\newblock A new model of information content based on concepts topology for
  measuring semantic similarity in wordnet 1.
\newblock \emph{International Journal of Grid and Distributed Computing},
  5\penalty0 (3):\penalty0 81--94, 2012.

\bibitem[Mertens and Recker(2020)]{mertens2019new}
Willem Mertens and Jan Recker.
\newblock New guidelines for null hypothesis significance testing in
  hypothetico-deductive is research.
\newblock \emph{Journal of the Association for Information Systems},
  21\penalty0 (4), 2020.
\newblock \doi{10.17705/1jais.00629}.
\newblock URL \url{https://aisel.aisnet.org/jais/vol21/iss4/1}.

\bibitem[Mika(2007)]{MIKA20075}
Peter Mika.
\newblock Ontologies are us: A unified model of social networks and semantics.
\newblock \emph{Journal of Web Semantics}, 5\penalty0 (1):\penalty0 5--15,
  2007.
\newblock ISSN 1570-8268.
\newblock \doi{10.1016/j.websem.2006.11.002}.
\newblock URL
  \url{https://www.sciencedirect.com/science/article/pii/S1570826806000552}.
\newblock Selected Papers from the International Semantic Web Conference.

\bibitem[Miller and Charles(1991)]{Miller1991}
George~A. Miller and Walter~G. Charles.
\newblock {Contextual Correlates of Semantic Similarity}.
\newblock \emph{Language \& Cognitive Processes}, 6\penalty0 (1):\penalty0
  1--28, 1991.
\newblock ISSN 01690965.
\newblock \doi{10.1080/01690969108406936}.

\bibitem[Musen(2015)]{musen2015the}
Mark~A. Musen.
\newblock The {P}rot\'eg\'e project: a look back and a look forward.
\newblock \emph{AI matters. Association of Computing Machinery Specific
  Interest Group in Artificial Intelligence}, 1\penalty0 (4):\penalty0 4--12,
  2015.

\bibitem[Pernet(2016)]{pernet2017}
Cyril Pernet.
\newblock Null hypothesis significance testing: A guide to commonly
  misunderstood concepts and recommendations for good practice.
\newblock \emph{F1000Research}, 4\penalty0 (621), 2016.
\newblock \doi{10.12688/f1000research.6963.5}.

\bibitem[Prudhomme et~al.(2020)Prudhomme, Homburg, Ponciano, Boochs, Cruz, and
  Roxin]{prudhomme2020interpretation}
Claire Prudhomme, Timo Homburg, Jean-Jacques Ponciano, Frank Boochs, Christophe
  Cruz, and Ana-Maria Roxin.
\newblock Interpretation and automatic integration of geospatial data into the
  semantic web.
\newblock \emph{Computing}, 102\penalty0 (2):\penalty0 365--391, 2020.
\newblock \doi{10.1007/s00607-019-00701-y}.

\bibitem[Resnik(1995)]{Resnik1995}
Philip Resnik.
\newblock Using information content to evaluate semantic similarity in a
  taxonomy.
\newblock In \emph{Proceedings of the 14th International Joint Conference on
  Artificial Intelligence - Volume 1}, IJCAI'95, pages 448--453, San Francisco,
  CA, USA, 1995. Morgan Kaufmann Publishers Inc.
\newblock ISBN 1-55860-363-8, 978-1-558-60363-9.
\newblock URL \url{http://dl.acm.org/citation.cfm?id=1625855.1625914}.

\bibitem[Rezaei and Fr{\"a}nti(2014)]{Reza2014}
Mohammad Rezaei and Pasi Fr{\"a}nti.
\newblock Matching similarity for keyword-based clustering.
\newblock In Pasi Fr{\"a}nti, Gavin Brown, Marco Loog, Francisco Escolano, and
  Marcello Pelillo, editors, \emph{Structural, Syntactic, and Statistical
  Pattern Recognition. S+SSPR 2014. Lecture Notes in Computer Science, vol
  8621}, pages 193--202, Berlin, Heidelberg, 2014. Springer Berlin Heidelberg.
\newblock ISBN 978-3-662-44415-3.
\newblock \doi{10.1007/978-3-662-44415-3_20}.

\bibitem[Rubenstein and Goodenough(1965)]{Rubenstein1965}
Herbert Rubenstein and John~B. Goodenough.
\newblock Contextual correlates of synonymy.
\newblock \emph{Communications of the ACM}, 8\penalty0 (10):\penalty0 627--633,
  1965.
\newblock ISSN 0001-0782.
\newblock \doi{10.1145/365628.365657}.

\bibitem[Sammut and Webb(2010)]{tfidf}
Claude Sammut and Geoffrey~I. Webb, editors.
\newblock \emph{{TF--IDF}}, pages 986--987.
\newblock Springer US, Boston, MA, 2010.
\newblock ISBN 978-0-387-30164-8.
\newblock \doi{10.1007/978-0-387-30164-8_832}.

\bibitem[S\'{a}nchez et~al.(2011)S\'{a}nchez, Batet, and Isern]{Sanchez2011}
David S\'{a}nchez, Montserrat Batet, and David Isern.
\newblock Ontology-based information content computation.
\newblock \emph{Knowledge-Based Systems}, 24\penalty0 (2):\penalty0 297--303,
  March 2011.
\newblock ISSN 0950-7051.
\newblock \doi{10.1016/j.knosys.2010.10.001}.

\bibitem[Schrijver(2003)]{Schri2003}
Alexander Schrijver.
\newblock \emph{{Combinatorial Optimization: Polyhedra and Efficiency}}.
\newblock Springer, 2003.
\newblock ISBN 0937-5511.

\bibitem[Seco et~al.(2004)Seco, Veale, and Hayes]{Seco2004}
Nuno Seco, Tony Veale, and Jer Hayes.
\newblock An intrinsic information content metric for semantic similarity in
  wordnet.
\newblock In \emph{Proceedings of the 16th European Conference on Artificial
  Intelligence}, ECAI'04, pages 1089--1090, Amsterdam, The Netherlands, 2004.
  IOS Press.
\newblock ISBN 978-1-58603-452-8.
\newblock URL \url{http://dl.acm.org/citation.cfm?id=3000001.3000272}.

\bibitem[Shajalal and Aono(2019)]{Shajalal2019}
Md. Shajalal and Masaki Aono.
\newblock Semantic textual similarity between sentences using bilingual word
  semantics.
\newblock \emph{Progress in Artificial Intelligence}, 8:\penalty0 263--272,
  2019.
\newblock ISSN 2192-6352.
\newblock \doi{10.1007/s13748-019-00180-4}.
\newblock URL
  \url{https://link.springer.com/article/10.1007/s13748-019-00180-4}.

\bibitem[Sj{\o}eberg et~al.(2005)Sj{\o}eberg, Hannay, Hansen, Kampenes,
  Karahasanovic, Liborg, and Rekdal]{1514443}
Dag I.~K. Sj{\o}eberg, Jo~Erskine Hannay, Ove Hansen, Vigdis~By Kampenes, Amela
  Karahasanovic, Nils~Kristian Liborg, and Anette~C. Rekdal.
\newblock A survey of controlled experiments in software engineering.
\newblock \emph{IEEE Transactions on Software Engineering}, 31\penalty0
  (9):\penalty0 733--753, Sept 2005.
\newblock ISSN 0098-5589.
\newblock \doi{10.1109/TSE.2005.97}.

\bibitem[Szumlanski et~al.(2013)Szumlanski, Gomez, and Sims]{Szumlanski2013}
Sean~R. Szumlanski, Fernando Gomez, and Valerie~K. Sims.
\newblock A new set of norms for semantic relatedness measures.
\newblock In \emph{ACL (2)}, pages 890--895. The Association for Computer
  Linguistics, 2013.
\newblock ISBN 978-1-937284-51-0.

\bibitem[Taglino and {De Nicola}(2023)]{https://doi.org/10.17632/r4vbkhgxx3.2}
Francesco Taglino and Antonio {De Nicola}.
\newblock {ACM} dataset for experimental assessment of semantic similarity
  methods, 2023.
\newblock URL \url{https://data.mendeley.com/datasets/r4vbkhgxx3/2}.

\bibitem[Tien et~al.(2019)Tien, Le, Tomohiro, and Tatsuya]{Huy2019}
Huy~Nguyen Tien, Minh~Nguyen Le, Yamasaki Tomohiro, and Izuha Tatsuya.
\newblock Sentence modeling via multiple word embeddings and multi-level
  comparison for semantic textual similarity.
\newblock \emph{Information Processing \& Management}, 56\penalty0 (6), 2019.
\newblock \doi{10.1016/j.ipm.2019.102090}.
\newblock URL \url{https://doi.org/10.1016/j.ipm.2019.102090}.

\bibitem[Toch et~al.(2011)Toch, Reinhartz-Berger, and Dori]{TOCH201116}
Eran Toch, Iris Reinhartz-Berger, and Dov Dori.
\newblock Humans, semantic services and similarity: A user study of semantic
  web services matching and composition.
\newblock \emph{Journal of Web Semantics}, 9\penalty0 (1):\penalty0 16--28,
  2011.
\newblock ISSN 1570-8268.
\newblock \doi{10.1016/j.websem.2010.10.002}.
\newblock URL
  \url{https://www.sciencedirect.com/science/article/pii/S1570826810000818}.

\bibitem[Tversky(1977)]{Tversky1977}
A.~Tversky.
\newblock Features of similarity.
\newblock \emph{Psychological Review}, 84\penalty0 (4):\penalty0 327--352,
  1977.
\newblock \doi{10.1037/0033-295X.84.4.327}.

\bibitem[Wu and Palmer(1994)]{WuPalmer}
Zhibiao Wu and Martha Palmer.
\newblock Verb semantics and lexical selection.
\newblock 32nd Annual meeting of the Associations for Computational
  Linguistics, pages 133--138, 1994.

\bibitem[Yang et~al.(2020)Yang, Wei, Guo, and Tan]{YANG2020100578}
Shuo Yang, Ran Wei, Jingzhi Guo, and Hengliang Tan.
\newblock Chinese semantic document classification based on strategies of
  semantic similarity computation and correlation analysis.
\newblock \emph{Journal of Web Semantics}, 63:\penalty0 100578, 2020.
\newblock ISSN 1570-8268.
\newblock \doi{10.1016/j.websem.2020.100578}.
\newblock URL
  \url{https://www.sciencedirect.com/science/article/pii/S1570826820300238}.

\bibitem[Zad et~al.(2021)Zad, Heidari, Hajibabaee, and Malekzadeh]{Samira2021}
Samira Zad, Maryam Heidari, Parisa Hajibabaee, and Masoud Malekzadeh.
\newblock A survey of deep learning methods on semantic similarity and sentence
  modeling.
\newblock In \emph{2021 IEEE 12th Annual Information Technology, Electronics
  and Mobile Communication Conference (IEMCON)}, pages 466--472, 2021.
\newblock \doi{10.1109/IEMCON53756.2021.9623078}.

\bibitem[Zhu and Iglesias(2017)]{Zhu2017}
Ganggao Zhu and Carlos~Angel Iglesias.
\newblock {Computing Semantic Similarity of Concepts in Knowledge Graphs}.
\newblock \emph{IEEE Transactions on Knowledge and Data Engineering},
  29:\penalty0 72--85, 2017.

\end{thebibliography}


\newpage

\section*{Annex A}

\begin{table*}[!h] 
	\centering 
	\caption{Number of papers indexed by SCOPUS  with ``semantic similarity'' grouped by application sectors.} 
	\label{tab:SectorsOccurrences} 
	\vspace*{0.1in} 
	\scriptsize 
	\begin{tabular}{| l | c | c |} 
		\hline 
		\textbf{Sectors} & \textbf{Number of papers} 
		\tabularnewline 
		\hline 
		Computer Science & 3,680   \tabularnewline 
		\hline 
		Mathematics & 1,116   \tabularnewline 
		\hline 
		Engineering & 980   \tabularnewline 
		\hline 
		Social Sciences & 429   \tabularnewline 
		\hline 
		Decision Sciences & 247   \tabularnewline 
		\hline 
		Arts and Humanities & 241   \tabularnewline 
		\hline 
		Biochemistry, Genetics and Molecular Biology & 180   \tabularnewline 
		\hline 
		Medicine & 171   \tabularnewline 
		\hline 
		Business, Management and Accounting & 140   \tabularnewline 		
		\hline 
		Materials Science & 58   \tabularnewline 
		\hline 
		Physics and Astronomy & 58   \tabularnewline 
		\hline 
		Psychology &  55  \tabularnewline 
		\hline 
		Earth and Plenatary Sciences &  45  \tabularnewline 
		\hline 
		Neuroscience &    35 \tabularnewline 
		\hline 
		Health Professions & 30   \tabularnewline 
		\hline 
		Chemical Engineering & 23   \tabularnewline 
		\hline 
		Environmental Science &  22  \tabularnewline
		\hline 
		Economics, Econometrics and Finance & 20   \tabularnewline 
		\hline 
		Chemistry & 18   \tabularnewline 
		\hline 
		Agricultural and Biological Sciences & 17    \tabularnewline 
		\hline 
		Energy &  16  \tabularnewline 
		\hline 
		Pharmacology, Toxicology and Pharmaceutics &  9 \tabularnewline 
		\hline 
		Immunology and Microbiology &  5  \tabularnewline 
		\hline 
		Nursing &  3  \tabularnewline 
		\hline 
		Veterinary & 1    \tabularnewline 
		\hline 
		Undefined &  1  \tabularnewline 
		\hline 
		Multidisciplinary &  32  \tabularnewline 
		\hline 
	\end{tabular} 
	\centering 
\end{table*}
 
\pagebreak

\section*{Annex B}

\begin{figure}[h]
	\centering
	\includegraphics[width=50mm]{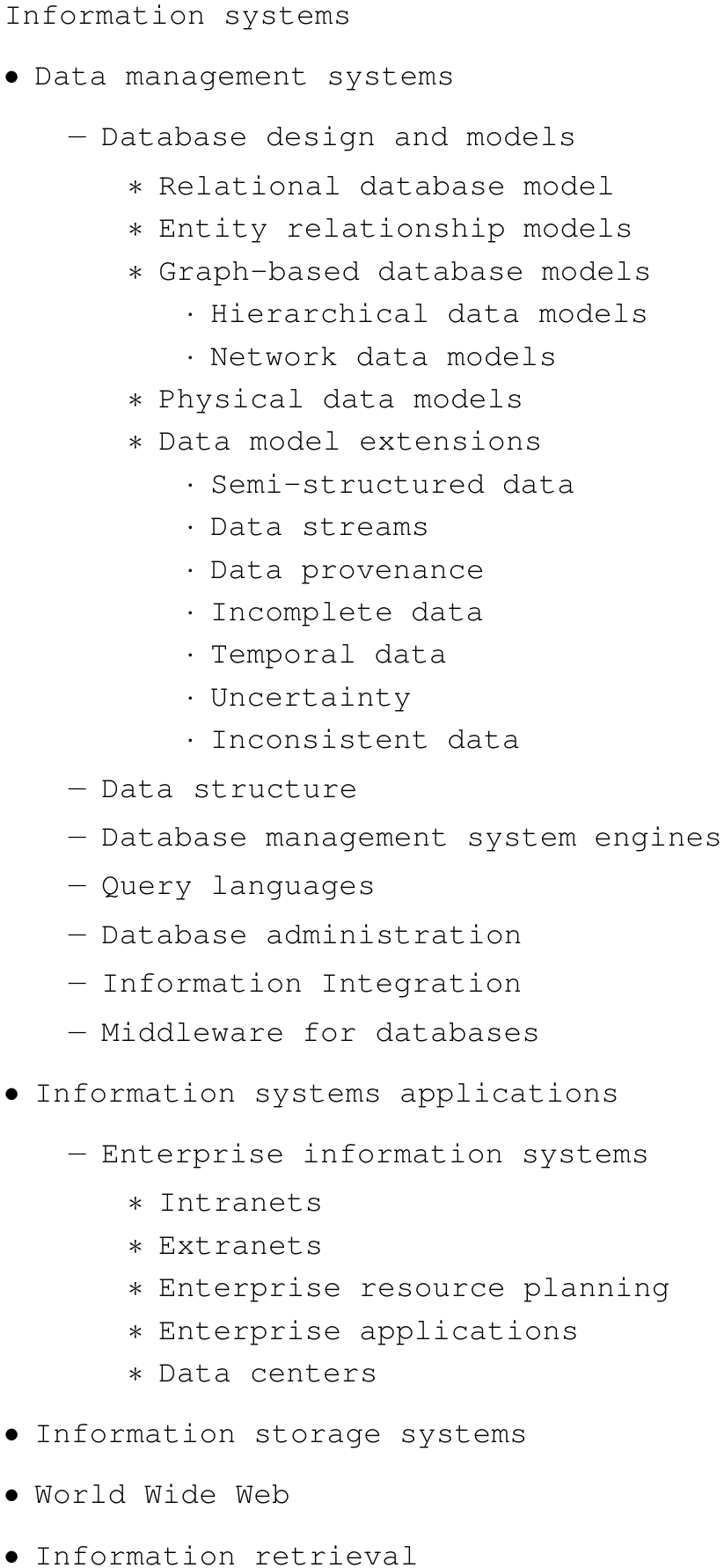}
	\caption{An excerpt of the ACM-CCS keywords.}
	\label{fig:acm_excerpt}
\end{figure}


\pagebreak

\section*{Annex C}

\begin{table*}[h]
\centering
\caption{Formulas of the set-theoretic similarity methods selected for comparison.}
\label{tab:SetTheoreticMethods} 
\vspace*{0.1in}
\scriptsize

{\renewcommand\arraystretch{1.15}
\begin{tabular}{| l | l |}
\hline
 \multirow{1}{*}{\textbf{Method}} & \multirow{1}{*}{\raggedright \textbf{Formula}}  \tabularnewline 
\hline
\hline
 \multirow{3}{*}{Dice} & \multirow{3}{260pt}{\raggedright 
\begin{equation*}
sim(av_1, av_2) = \frac{2|av_1 \cap av_2|}{|av_1| + |av_2|} 
\end{equation*}} \tabularnewline 
  & \tabularnewline 
  & \tabularnewline 
 \hline
 
\multirow{3}{*}{Jaccard} & \multirow{3}{260pt}{\begin{equation*}
sim(av_1, av_2) = \frac{|av_1 \cap av_2|}{|av_1 \cup av_2|} 
\end{equation*}}  \tabularnewline 
  & \tabularnewline 
  & \tabularnewline 
 \hline

\multirow{3}{*}{Sigmoid} & \multirow{3}{260pt}{\begin{equation*}
sim(av_1, av_2) = \frac{e^{|av_1 \cap av_2|}-1}{(e^{|av_1 \cap av_2|}+1)(|av_1 -  av_2|+|av_2 -  av_1|+1)}
\end{equation*}}  \tabularnewline 
  & \tabularnewline 
  & \tabularnewline 
 \hline

\end{tabular}}

\centering
\end{table*}


\begin{table*}
\centering
\caption{Formulas of the taxonomy-based similarity methods selected for comparison.}
\label{tab:TaxonomyBasedMethods} 
\vspace*{0.1in}
\scriptsize

{\renewcommand\arraystretch{1.15}
\begin{tabular}{| l | l |}
\hline
 \multirow{1}{*}{\textbf{Method}} & \multirow{1}{*}{\raggedright \textbf{Formula}}  \tabularnewline 
\hline
\hline
 
\multirow{4}{*}{} & \multirow{4}{260pt}{\begin{equation*}
sim (av_1,av_2)= \frac{\displaystyle\sum_{c_i \in av_1}  max_{c_j \in av_2}(sim^{lch}(c_i,c_j)) \cdot IDF (c_i)}{\displaystyle\sum_{c_i \in av_1} IDF (c_i)}
\end{equation*}} \tabularnewline 
  &  \tabularnewline 
  & \tabularnewline 
  & \tabularnewline 

\multirow{1}{*}{WNSim} & \multirow{1}{*}  where  \tabularnewline

\multirow{1}{*}{} & \multirow{1}{*}  \text{}  $sim^{lch} (c_i,c_j)=-log  \frac{length(c_i,c_j)}{2 \cdot maxdepth}$ \tabularnewline


\multirow{1}{*}{\textit{}} & \multirow{1}{*}  \text{} $IDF$ is defined according to Eq. (\ref{eq:idf}) in Section \ref{sec:ExtensionalMethods} \tabularnewline

\multirow{1}{*}{\textit{}} & \multirow{1}{*}  \text{} $length(c_i, c_j)$ is length of the shortest path between $c_i$ and $c_j$ \tabularnewline


\multirow{1}{*}{\textit{}} & \multirow{1}{*} \text{} $maxdepth$ is the maximum depth of the taxonomy \tabularnewline


 \hline

\multirow{3}{*}{} & \multirow{3}{260pt}{\begin{equation*}
sim(av_1, av_2)=\frac{\displaystyle\sum_{c_i \in av_1}S(c_i,c_{j})}{|av_1|} 
\end{equation*}}  \tabularnewline 
  & \tabularnewline 
  & \tabularnewline 
  
 \multirow{1}{*}{} & \multirow{1}{*}  where    \tabularnewline

 \multirow{1}{*}{Rezaei \& Fr{\"a}nti} & \multirow{1}{*}  \text{}  $|av_1| \geq |av_2|$  \tabularnewline 

 \multirow{1}{*}{} & \multirow{1}{*}  \text{}  $c_j \in av_2$ is the concept paired with $c_i$ according to Section \ref{sec:SimSetsOfConcepts}  \tabularnewline 

 \multirow{1}{*}{} & \multirow{1}{*}  \text{}  $S(c_i,c_j)=\frac{2 \cdot depth(lcs(c_i,c_j))}{depth(c_i)+depth(c_j)}$   \tabularnewline


  \multirow{1}{*}{} & \multirow{1}{*}  \text{}  $depth(c)$ is the length of the shortest path between $c$ and the root of the taxonomy  \tabularnewline 

 \multirow{1}{*}{} & \multirow{1}{*}  \text{} $lcs(c_i, c_j)$ is the least common subsumer of $c_i$ and $c_j$ \tabularnewline

  
 \hline
 
\multirow{3}{*}{} & \multirow{3}{260pt}{\begin{equation*}
sim(av_1,av_2)=\frac{1}{|av_1|}\sum_{c_i \in av_1} max_{c_j \in av_2}S(c_i,c_j)
\end{equation*}}  \tabularnewline 
  & \tabularnewline 
  & \tabularnewline 
  
  \multirow{1}{*}{Haase et al.} & \multirow{1}{*}  where \tabularnewline
 
 \multirow{4}{*}{} & \multirow{4}{260pt}{
$$
S(c_i,c_j) = \left\{
\begin{array}{lll}
e^{-\alpha l}.\displaystyle \frac{e^{\beta h}-e^{-\beta h}}{e^{\beta h}+e^{-\beta h}}& & \text{if } c_i \neq c_j \\
1 & &\mbox{otherwise}
\end{array}
\right.
$$
}  \tabularnewline 
  & \tabularnewline 
  & \tabularnewline 
  & \tabularnewline

\multirow{1}{*}{} & \multirow{1}{*} \text{} $\alpha = 0.2$   \tabularnewline 
\multirow{1}{*}{} & \multirow{1}{*} \text{} $\beta = 0.6$   \tabularnewline 
\multirow{1}{*}{} & \multirow{1}{*} \text{} $h=depth(lcs(c_i,c_j))$   \tabularnewline 
\multirow{1}{*}{} & \multirow{1}{*} \text{} $l=length(c_i, c_j)$  \tabularnewline 

\hline

\end{tabular}}

\centering
\end{table*}





\end{document}